\documentclass[10pt,twocolumn,letterpaper]{article}
\usepackage[accsupp]{axessibility}
\usepackage{cvpr}
\usepackage{multirow}
\usepackage{balance}
\usepackage[dvipsnames]{xcolor}

\definecolor{cvprblue}{rgb}{0.21,0.49,0.74}
\usepackage[pagebackref,breaklinks,colorlinks,citecolor=cvprblue]{hyperref}

\def\paperID{****} 
\def\confName{CVPR}
\def\confYear{2024}

\title{Monocular Identity-Conditioned Facial Reflectance Reconstruction}

\author{Xingyu Ren\quad Jiankang Deng\textsuperscript{*}\quad Yuhao Cheng\quad Jia Guo\quad Chao Ma\textsuperscript{*} \\ Yichao Yan\quad Wenhan Zhu\quad Xiaokang Yang
\and
MoE Key Lab of Artificial Intelligence, AI Institute, Shanghai Jiao Tong University\\
{\tt\small \{rxy\_sjtu,chengyuhao,chaoma,yanyichao,zhuwenhan823,xkyang\}@sjtu.edu.cn}\\
\tt\small \{jiankangdeng,guojia\}@gmail.com}

\begin{document}
\twocolumn[{%
\renewcommand\twocolumn[1][]{#1}%
\maketitle
\begin{center}
    \centering
    \captionsetup{type=figure}
    \includegraphics[width=\linewidth]{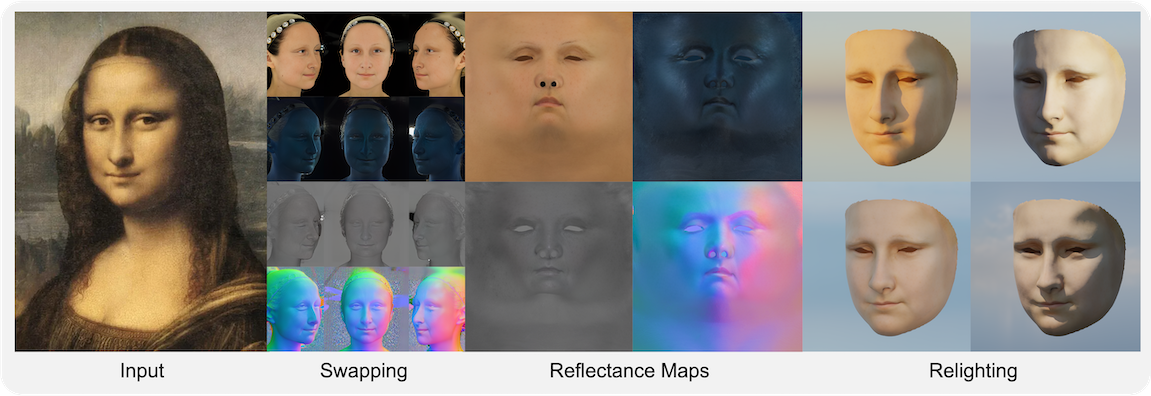}
    \caption{We present ID2Reflectance, a high-quality, identity-conditioned reflectance reconstruction method. ID2Reflectance learns multi-domain face codebooks by using limited captured data and generates multi-view domain-specific reflectance images guided by facial identity. Our approach greatly reduces the dependency on captured data and generates high-fidelity reflectance maps for realistic rendering.}
    \label{fig:teser}
\end{center}%
}]
\footnotetext{\textsuperscript{*}Corresponding authors}
\begin{abstract}
Recent 3D face reconstruction methods have made remarkable advancements, yet there remain huge challenges in monocular high-quality facial reflectance reconstruction.
Existing methods rely on a large amount of light-stage captured data to learn facial reflectance models. However, the lack of subject diversity poses challenges in achieving good generalization and widespread applicability.
In this paper, we learn the reflectance prior in image space rather than UV space and present a framework named ID2Reflectance. Our framework can directly estimate the reflectance maps of a single image while using limited reflectance data for training.
Our key insight is that reflectance data shares facial structures with RGB faces, which enables obtaining expressive facial prior from inexpensive RGB data thus reducing the dependency on reflectance data.
We first learn a high-quality prior for facial reflectance. Specifically, we pretrain multi-domain facial feature codebooks and design a codebook fusion method to align the reflectance and RGB domains.
Then, we propose an identity-conditioned swapping module that injects facial identity from the target image into the pre-trained autoencoder to modify the identity of the source reflectance image. Finally, we stitch multi-view swapped reflectance images to obtain renderable assets.
Extensive experiments demonstrate that our method exhibits excellent generalization capability and achieves state-of-the-art facial reflectance reconstruction results for in-the-wild faces.
Our project page is \href{https://xingyuren.github.io/id2reflectance/}{https://xingyuren.github.io/id2reflectance/}.
\end{abstract}
\section{Introduction}
\label{sec:intro}

Facial reflectance reconstruction aims at predicting reflectance components (\eg, diffuse and specular albedo) and high-frequency normals of the input in-the-wild face image. The recovered 3D faces can be realistically rendered in arbitrary illumination environments. Facial reflectance reconstruction is one of the fundamental problems in computer vision and graphics, with applications ranging from avatar creation~\cite{zhang2023dreamface}, telecommunication~\cite{ma2021pixel}, video games, films, and interactive AR/VR.

To achieve a realistic facial reflectance model, it is necessary to collect a large-scale and high-quality reflectance dataset from various individuals.
However, capturing high-fidelity facial reflectance data is costly and time-consuming, requiring specialized scanning equipment (\eg, Light Stage~\cite{debevec2000acquiring,ghosh2011multiview}) and skilled artists for post-processing. Therefore, recent efforts~\cite{smith2020AlbedoMM,Lattas20,li2020learning} only manage to collect facial reflectance datasets with less than three hundred subjects. As a result, these reflectance models~\cite{smith2020AlbedoMM,Lattas20,li2020learning,luo2021normalized} can not generalize very well across diverse real-world identities. To this end, AvatarMe++~\cite{lattas2021avatarme++} employs an image-to-image translation network~\cite{isola2017image} to synthesize diffuse, specular, and normal maps from large-scale facial texture maps~\cite{booth20163d}. Based on the complete pairs of facial reflectance, FitMe~\cite{lattas2023fitme} and Relightify~\cite{papantoniou2023relightify} achieve good performance in facial reflectance reconstruction by using StyleGAN~\cite{Karras2019stylegan2} and latent diffusion model~\cite{rombach2022stablediffusion} to learn facial reflectance prior. Nevertheless, none of the facial reflectance data is released from these works~\cite{lattas2021avatarme++,lattas2023fitme,papantoniou2023relightify}.

\begin{figure*}[t]
    \centering
    \includegraphics[width=\linewidth]{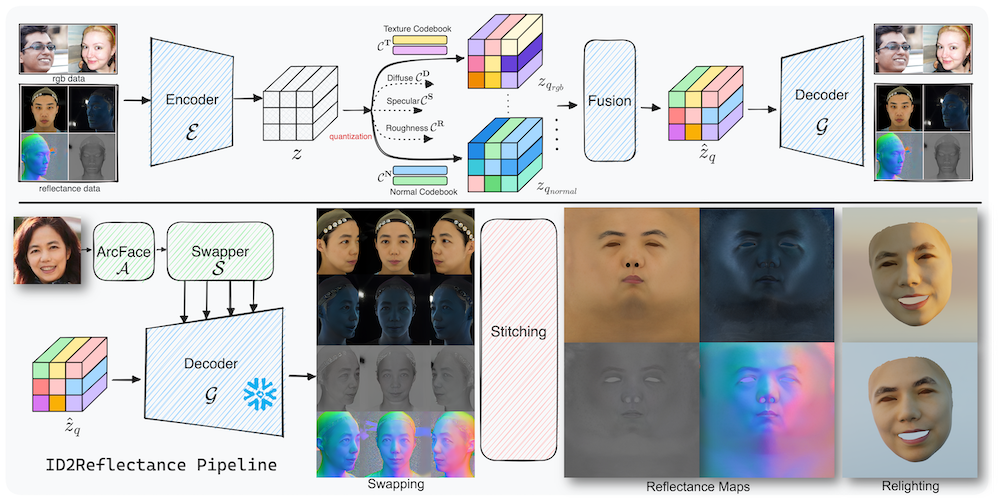}
    \caption{Overview of the proposed method. Our core insight is to build a facial reflectance prior in image space by using limited captures and to recover the reflectance maps for any unconstrained face. We first train multi-domain facial codebooks using a large amount of RGB data and limited reflectance data. Then, given an input unconstrained face, we extract the identity feature from the pre-trained ArcFace~\cite{deng2019arcface} model.
    This feature is fed into the swapper module, which guides the decoder to perform identity injection for all domains. We finally stitch three-view identity-conditioned reflectance images to acquire high-quality rendering assets and renderable 3D faces.}
    \label{fig:pipeline}
\end{figure*}

Since the current facial reflectance models~\cite{smith2020AlbedoMM,Lattas20,li2020learning} are mainly trained in the UV space, a considerable amount of facial reflectance data is needed to learn facial structure, chromaticity, complex details, and other features from scratch. To this end, we train the facial reflectance model with limited light stage captures in the image space instead of the UV space, thus we can take advantage of large-scale, high-quality, and diverse RGB images (\eg, FFHQ~\cite{karras2019style}). As shown in Fig.~\ref{fig:pipeline}, facial reflectance data (\eg, diffuse albedo, specular albedo, roughness, and surface normal) share the same facial structure as the RGB faces in the image space. Through joint training, basic facial structure priors can be learned from inexpensive RGB data, leading to a significant reduction in the necessity of reflectance data.

To learn a joint facial reflectance and RGB model, we combine multi-domain data to train VQGAN~\cite{van2017vqvae,esser2021taming,zhou2022towards,liu2023learning}, which employs discrete generative priors, in terms of codebooks, for high-quality image synthesis. However, it is difficult to rely on a single codebook to reconstruct vastly different facial reflectance images as obvious artifacts can be observed in the reconstructions (Fig.~\ref{fig:codebook}). To this end, we design a multi-domain codebook learning scheme and each codebook represents a domain-specific discretization of the latent space. For an input image, the final latent representation is a weighted combination derived from these codebooks, resulting in a more expressive and robust representation.

After we train the facial reflectance model, we further apply it for unconstrained facial reflectance reconstruction.
Inspired by ID2Albedo~\cite{id2albedo}, we employ identity-conditioned reflectance prediction instead of employing the iterative fitting~\cite{lattas2023fitme} or conditional inpainting~\cite{papantoniou2023relightify}. 
To inject identity information from the target face into the pre-trained quantized autoencoder, we propose an identity integration module by using AdaIN~\cite{huang2017arbitrary} and train identity swapping only in the RGB domain. 
As the facial reflectance domain and RGB domain are aligned in our VQGAN model, the identity-swapping capacity learned from the RGB domain can be automatically transferred to the facial reflectance domain. To obtain the complete reflectance maps in UV space, we synthesize three-view identity-conditioned reflectance images in the wrapped space and finally stitch them together for realistic rendering, as illustrated in Fig.~\ref{fig:teser}.

In summary, we make the following contributions:
\begin{itemize}[noitemsep,nolistsep]
\item We propose a novel facial reflectance reconstruction framework that utilizes multi-domain codebooks to align the facial reflectance domain with the RGB domain to significantly reduce the requirement of captured data.
\item We propose a lightweight face swapper module to inject the identity feature into the pre-trained decoder to achieve identity-conditioned facial reflectance generation.
\item Extensive experiments demonstrate that the proposed ID2Reflectance can predict high-fidelity facial reflectance from in-the-wild face images.
\end{itemize}
\section{Related Work}
\label{sec:2_related_works}

\noindent{\bf Facial Reflectance Reconstrucion.}
3D Morphable Models (3DMMs)~\cite{Blanz99,Egger2020} are typical approaches for face reconstruction from unconstrained images. The linear parametric face model constrains face reconstruction in a low-dimensional space by encoding facial shapes and textures with Principal Component Analysis (PCA), thus neglecting high-frequency facial details. 
To achieve high-fidelity facial representation, non-linear 3DMM methods~\cite{tran2018nonlinear,Tewari19,lee2020uncertainty,Feng2021} are introduced. These models are formulated as neural network decoders where the 3D faces are generated directly from latent vectors. To obtain high-quality facial texture, GANFit~\cite{Gecer19:ganfit} employs ProgressiveGAN~\cite{karras2017progressive} as the texture generator~\cite{deng2018uv,gecer2020synthesizing}. However, GANFit lacks relighting capabilities due to backed illumination in the texture.

To overcome the backed illumination problem, the Light Stage~\cite{debevec2000acquiring} is employed to capture high-quality facial reflectance data~\cite{smith2020AlbedoMM,li2020learning,Lattas20}. By using multiple gradient illuminations with polarization~\cite{ghosh2011multiview}, the diffuse and specular components of reflection can be separated.
Given reflectance data, face reconstruction is upgraded into facial reflectance reconstruction~\cite{Yamaguchi18,li2020learning,smith2020AlbedoMM,Lattas20,luo2021normalized,dib2021towards,lattas2023fitme,han2023learning,papantoniou2023relightify,zhang2023dreamface,bai2023ffhquv}.
AlbedoMM~\cite{smith2020AlbedoMM} first proposes 
a drop-in replacement to the 3DMM statistical albedo model with separate diffuse and specular albedo priors, but AlbedoMM is still based on a linear per-vertex color model. 
Since the capture process by the light stage is expensive and time-consuming, the identity number is usually limited to several hundreds~\cite{smith2020AlbedoMM,li2020learning,Lattas20}.
To this end, AvatarMe++~\cite{lattas2021avatarme++} synthesizes diffuse and specular colors for high-quality textures~\cite{booth20163d} by training an image-to-image translation network using limited lightstage data. Based on the data from AvatarMe++~\cite{lattas2021avatarme++}, FitMe~\cite{lattas2023fitme} proposes a BRDF generative network and a two-stage fitting method to predict facial reflectance for unconstrained images. Relightify~\cite{papantoniou2023relightify} utilizes a powerful diffusion model to infer diffuse, specular, and normal.

Even though the series of AvatarMe~\cite{Lattas20}, FitMe~\cite{lattas2023fitme}, and Relightify~\cite{papantoniou2023relightify} achieve good performance in facial reflectance reconstruction, neither the texture synthesized reflectance data nor the light stage captured reflectance data is released.
In this paper, we define the facial reflectance model in the image space instead of the widely-used UV space, thus we can take advantage of large-scale and high-quality RGB faces to learn facial structure priors. By training multi-domain aligned codebooks, we only require limited reflectance training data for facial reflectance reconstruction.

\noindent{\bf Face Swapping.}
The task of face swapping is to transfer the facial identity of the source image/video into the target image/video. Early works mainly utilize traditional image processing~\cite{bitouk2008faceswapping} and 3D morphable models~\cite{vetter19983dmm}. Recent methods~\cite{wang2021hififace,gao2021inforswap,zhu2021megafs,xu2022regionfs,xu2022styleswap,luo2022styleface,xu2022uniface,li20233dfaceswapping,shiohara2023blendface,ren2023reinforced} heavily rely on Generative Adversarial Networks~\cite{karras2019style,Karras2019stylegan2} and advanced face recognition models~\cite{deng2019arcface} to achieve photo-realistic and identity-preserved face swapping.  
However, all of these face-swapping methods are designed for the RGB domain and cannot be directly used when the target face is from the reflectance domain. 
In this paper, we first train multi-domain codebooks by using VQGAN~\cite{esser2021taming}.
Then, we design an identity injection module for the decoder by using AdaIN~\cite{huang2017arbitrary,li2020faceshifter,chen2020simswap} to train face-swapping in the RGB domain. As the facial reflectance codebooks and RGB face codebook are aligned during our multi-domain codebook learning, the swapping capacity in the RGB domain can be automatically transferred to the reflectance domain. Therefore, we can decode high-quality identity-conditioned facial reflectance when the input of the encoder is from the facial reflectance domain.

\section{Methodology}
This work aims to reconstruct high-quality reflectance maps for a single unconstrained face image.
To this end, we first train a high-quality facial reflectance model through a multi-domain codebook learning scheme (Sec.~\ref{sec:3-1}). 
Based on the pre-trained multi-domain codebooks and a pre-trained face recognition model~\cite{deng2019arcface}, we train face swapping in the RGB domain and automatically transfer the swapping capacity to the reflectance domain (Sec.~\ref{sec:3-2}).
As illustrated in Fig.~\ref{fig:pipeline}, we finally design a simple yet efficient inference framework to achieve monocular high-quality facial reflectance map reconstruction (Sec.~\ref{sec:3-3}).

\subsection{Codebook Learning}
\label{sec:3-1}

The main challenge in training expressive facial reflectance models is the absence of large-scale and high-quality reflectance maps scanned from diverse individuals. Existing reflectance models~\cite{Li20,Lattas20} are learned from scratch in the unwrapped facial UV space with limited captured data (several hundred subjects). As we can see from Fig.~\ref{fig:pipeline}, facial reflectance data (\eg, diffuse albedo $\mathbf{D}$, specular albedo $\mathbf{S}$, roughness $\mathbf{R}$ and surface normal $\mathbf{N}$) share the same facial structure prior as the common RGB faces in the image space. Since the face structure remains consistent in the image space regardless of different domains, we train a joint quantized autoencoder (\ie VQGAN~\cite{esser2021taming}) to learn the RGB and reflectance codebook simultaneously. Through joint training, the identity diversity in the large amount of inexpensive RGB data can be shared with the limited facial reflectance data. To achieve high-quality facial multi-modal codebooks, we adopt a two-stage training approach.

In the {\bf first stage}, we simply train the quantized autoencoder with a shared codebook using both high-quality facial RBG and reflectance data. For the single-channel data, we duplicate roughness into three-channel images and put specular albedo in the blue channel to simplify the training. 
Following VQGAN~\cite{esser2021taming}, we employ a quantized autoencoder~\cite{van2017vqvae} architecture which consists of a encoder~$\mathcal{E}$, a discrete codebook~$\mathcal{C}$, a decoder~$\mathcal{G}$, and a discriminator~$\mathcal{D}$. 
Given a high dimensional image $x \in \mathbb{R}^{{H} \times {W} \times 3}$, the encoder~$\mathcal{E}$ embeds the input image into the low dimensional code vector $z = \mathcal{E}(x) \in \mathbb{R}^{{h} \times {w} \times {d}}$, where ${d}=256$ is the dimension of the latent vector. 
Then, each grid vector in $z$ is replaced by the nearest vector from the learnable codebook $\mathcal{C} = \left \{ c_n \in \mathbb{R}^d \right \}^{N}_{n=1}$:
\begin{equation}
z^{(i,j)}_q = \mathbf{q}({z}^{(i,j)}) := \arg\min_{{c_n} \in \mathcal{C}} \Vert {{z}^{(i,j)}} - {c_n} \Vert,
\label{eq:vq}
\end{equation}
where $z_{q} \in \mathbb{R}^{{h} \times {w} \times {d}}$ is the quantized feature, $\mathbf{q}(\cdot)$ is a quantisation operation, and $N=1,024$ is the number of codes in the codebook. Taking the quantized representation $z_{q}$ as input, the decoder $\mathcal{G}$ can reconstruct the high-quality face image $\hat{x}=\mathcal{G}(\mathbf{q}(\mathcal{E}(x)))$).

To train the quantized autoencoder, we follow~\cite{esser2021taming} to employ three image-level losses: (1) photo reconstruction loss $\mathcal{L}_{photo} = \| \hat{x} - x \|_{1}$, (2) perceptual loss~\cite{zhang2018perceptual} $\mathcal{L}_{per} =  \sum_{l}\left \| \mathcal{V}_l(\hat{x}) - \mathcal{V}_l({x})\right \|_2^2$, where $l$ denotes the different level of a pre-trained VGG~\cite{simonyan2014very} model $\mathcal{V}$, and (3) adversarial loss~\cite{goodfellow2014generative} $\mathcal{L}_{adv1} = \log \mathcal{D}(x) + \log(1-\mathcal{D}(\hat{x}))$. As the quantization operation in Eq.~\ref{eq:vq} is non-differentiable, VQGAN~\cite{esser2021taming} simply copies the gradients from the decoder to the encoder. The intermediate code-level loss is:
\begin{equation}
\mathcal{L}_{code} = \left \| \mathrm{sg}\left [ \mathcal{E}(x) \right ] - {z_q} \right \|_{2}^{2} + \beta \left \| \mathrm{sg}\left [ {z_q} \right ] - \mathcal{E}(x) \right \|_{2}^{2},
\label{eq:code}
\end{equation}
where $\mathrm{sg}=\left [ \cdot \right ]$ denotes the stop-gradient operation and $\beta=0.25$ is controlling the update frequency of the codebook.

With the above image-level and code-level losses, we summarize the training objective as:
\begin{equation}
\mathcal{L}_{1} = \mathcal{L}_{photo} + \eta_{1}\mathcal{L}_{per} + \eta_{2}\mathcal{L}_{adv1} + \eta_{3}\mathcal{L}_{code},
\label{eq:total1}
\end{equation}
where the loss weights $\eta_{1}$, $\eta_{2}$, and $\eta_{3}$ are set as $1.5$, $0.2$ and $1.0$, respectively. After training, the shared codebook~$\mathcal{C}$ presents the facial features containing context-rich structures and details. However, a single codebook makes it hard to reconstruct vastly different facial reflectance images, and the reconstruction results sometimes contain artifacts (Fig.~\ref{fig:codebook}).

\begin{figure}
\centering
\includegraphics[width=\linewidth]{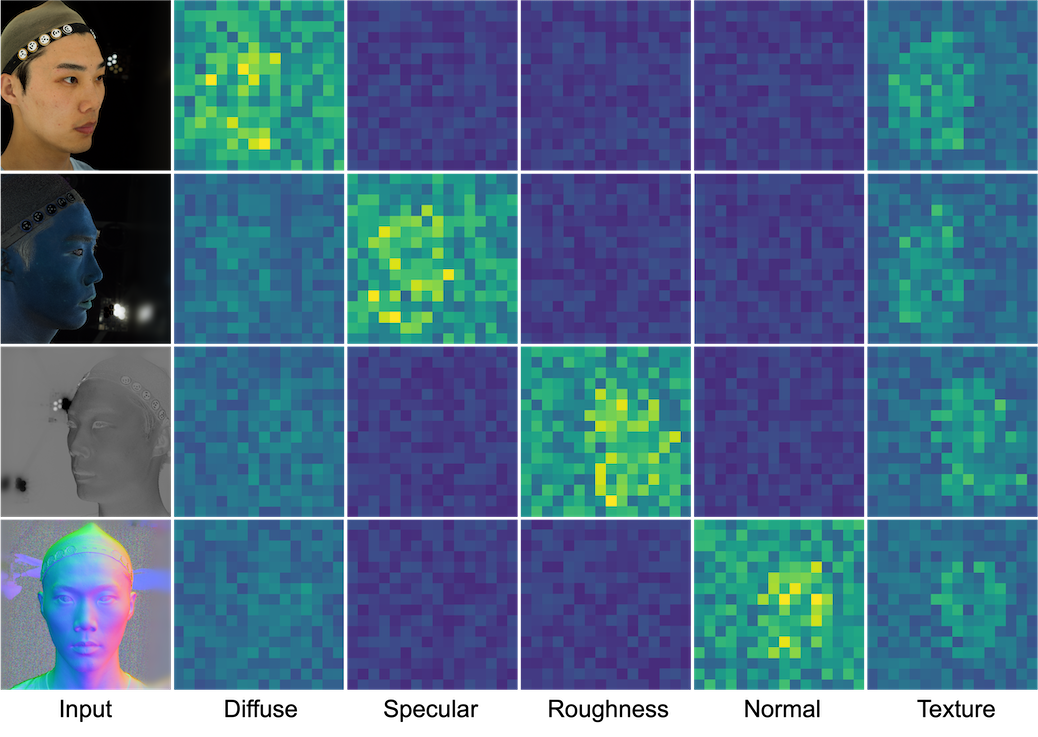}
\caption{Visualization of codebook fusion weights. 
Our method uses multiple basis codebooks (especially the RGB texture codebook) for discrete representations, indicating the cross-domain correlation learned by our model.}
\label{fig:fusionweight}
\end{figure}

In the {\bf second stage}, we train domain-specific codebooks to further improve facial reflectance reconstruction. More specifically, we fix the encoder and decoder, and fine-tune the codebook $\mathcal{C}$ learned from the first stage by separately using the reflectance data and the RGB data. Therefore, we obtain another five codebooks, \ie diffuse albedo codebook $\mathcal{C}^\mathbf{D}$, specular albedo codebook 
$\mathcal{C}^\mathbf{S}$, roughness codebook $\mathcal{C}^\mathbf{R}$, surface normal codebook $\mathcal{C}^\mathbf{N}$, and texture codebook $\mathcal{C}^\mathbf{T}$. 

For a given input $x$, five quantized representations $\left \{ z_{qk} \in \mathbb{R}^{h\times w\times d}\right \}^K_{k=1}$ can be generated by quantizing the latent code $z$ with the five domain-specific codebooks $\left \{  \mathcal{C}^\mathbf{D}, \mathcal{C}^\mathbf{S}, \mathcal{C}^\mathbf{R}, \mathcal{C}^\mathbf{N}, \mathcal{C}^\mathbf{T}\right \}$. To combine these five discrete representations $z_{q1,\cdots, K}$, we further train the swin transformer blocks~\cite{liu2021swin} as the fusion weight prediction module, which takes $z \in \mathbb{R}^{{h} \times {w} \times {d}}$ as input and outputs a weight map $w\in \mathbb{R}^{h\times w\times K}$. During the training of the fusion module, the encode, the decoder, and all codebooks are fixed. The code fusion can be expressed by
\begin{equation}
\hat{z}_q=\mathcal{W}(z_{q1,\cdots, K})=\sum_{k=1}^{K}w_kz_{qk},
\label{eq:fusion}
\end{equation}
where $\mathcal{W}(\cdot)$ denotes the code fusion operation. By decoding the fused code $\hat{z}_q$, we can get the reconstruction data $\hat{x}=\mathcal{G}(\hat{z}_q)$. As shown in Fig.~\ref{fig:fusionweight}, the reflectance data (\eg, normal) uses multiple basis codebooks (\eg, normal and RGB texture codebooks) for discrete representations, which indicates the alignment behavior between the reflectance domain and RGB domain.

\subsection{Identity Swapping}
\label{sec:3-2}
\begin{figure}
\centering
\includegraphics[width=\linewidth]{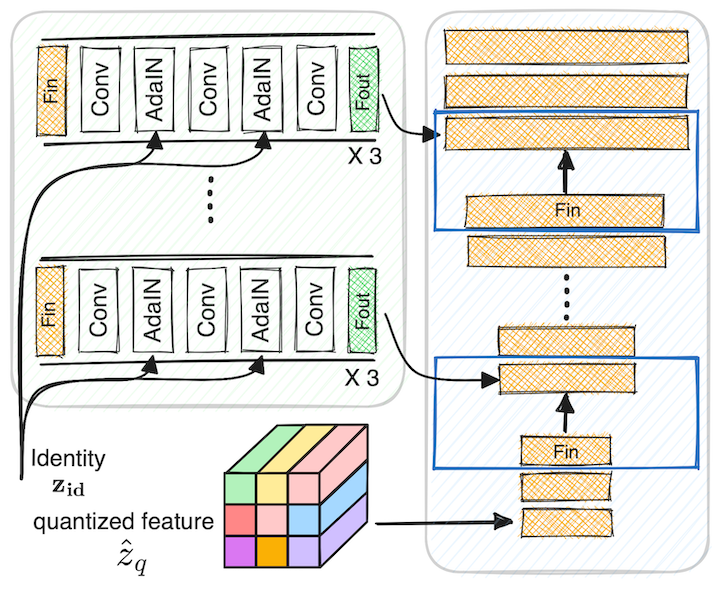}
\caption{Detailed architecture of our swapper module. The yellow boxes represent the original multi-scale features from the decoder $\mathcal{G}$, and the green boxes represent the residual features generated by each identity injection branch. We use the small-scale feature map as input to generate identity-conditioned residual features, which will be added to the up-sampled feature map.}
\label{fig:swapper}
\end{figure}

Since the facial reflectance domain and RGB domain are aligned during our multi-domain codebooks learning in Sec.~\ref{sec:3-1},
we explore training identity swapping~\cite{li2020faceshifter} in the RGB domain and automatically transferring the identity-swapping capacity to the facial reflectance domain. Specifically, we propose an identity-conditioned swapper module (denoted as~$\mathcal{S}$) and integrate it with the above-learned quantized autoencoder.

As illustrated in Fig.~\ref{fig:swapper}, our swapper module consists of serval parallel branches, each containing three identity injection blocks. 
Each block consists of convolution and AdaIN~\cite{huang2017arbitrary} operators and uses the Leaky ReLU as the activation function. 
We insert the swapper branch into each upsampling stage of the decoder $\mathcal{G}$. 
Specifically, we take the small-scale feature map as the input and use identity as the condition to generate a residual feature, which is finally incorporated into the up-sampled feature map. 
The identity embedding network is a ResNet-100 model trained on the large-scale WebFace dataset~\cite{zhu2021webface260m} using the ArcFace loss~\cite{deng2019arcface}. 
By using this pre-trained ArcFace model, we can extract face identity features that are robust to changes in illumination, pose, and occlusion. 
For identity embedding integration, it is achieved by using AdaIN~\cite{huang2017arbitrary} as follows:
\begin{equation}
AdaIN(f,z_{id})=\sigma_{z_{id}}\frac{f-\mu_f}{\sigma_f}+\mu_{z_{id}},
\label{eq:adain}
\end{equation}
where $z_{id}\in\mathbb{R}^{1\times512}$ is the identity embedding,
$\mu_f$ and $\sigma_f$ are the channel-wise mean and standard deviation of the input feature $f$, and
$\sigma_{z_{id}}$ and $\mu_{z_{id}}$ are two modulation parameters generated from $z_{id}$ through FC layers.

To train our swapper module, we employ the RGB face recognition dataset~\cite{cao2018vggface2}.
During training, we fix the pre-trained encoder, codebooks, and decoder. 
We employ the identity loss function, which is the cosine distance between the input image $\mathbf{I_{id}}$ and the decoded face $\hat{x}$:
\begin{equation}
\mathcal{L}_{id} =  1 - \frac{\mathcal{A}(\mathbf{I_{id}})\mathcal{A}({\hat{x}})}{\|\mathcal{A}(\mathbf{I_{id}})\|_2\cdot \|\mathcal{A}({\hat{x}})\|_2},
\label{eq:identityloss}
\end{equation}
where $\mathcal{A}$ is the pre-trained ArcFace model. Besides, we utilize a pyramid discriminator by referring to projected GAN~\cite{Sauer2021ProjectedGAN}.
\begin{equation}
\begin{aligned}
\mathcal{L}_{adv2} = \min_\mathcal{G} \max_\mathcal{D}  &\sum_{l \in L} \Big ( E_{x} [\log \mathcal{D}_l( \mathcal{F}_l(x))] \\
& +  E_{\hat z_q}[ log( 1- \mathcal{D}_l(\mathcal{F}_l(\mathcal{G}(\hat z_q))))] \Big),
\end{aligned}
\label{eq:projectedGAN}
\end{equation}
where $l \in \{1,\cdots, L\}, L=4$ indicates different feature levels, $\mathcal{F}$ is a fixed ImageNet model mapping the high-resolution image $x$ or decoded image $\mathcal{G}(\hat z_q)$ into four-scale feature pyramids, and $\mathcal{D}_l$ is the corresponding discriminator applied to each feature level. The complete training objective for face swapping is as follows:
\begin{equation}
\mathcal{L}_{2} = \mathcal{L}_{id} + \lambda_{1}\emph{1}\mathcal{L}_{photo} + \lambda_{2}\mathcal{L}_{lpips} + \lambda_{3}\mathcal{L}_{adv2},
\label{eq:swapping}
\end{equation}
where $\emph{1}$ is the indicator function which is $1$ when the input face $x$ and the target face $\mathbf{I_{id}}$ are from the same identity and $0$ otherwise. $\mathcal{L}_{lpips}$ denotes the LPIPS loss~\cite{zhang2018perceptual}.
The loss weights $\lambda_{1}$, $\lambda_{2}$, $\lambda_{3}$ are set as $1.5$, $0.1$ and $0.1$, respectively.
Compared to general face-swapping networks~\cite{li2020faceshifter,chen2020simswap}, the proposed codebook-based identity swapping explicitly decouples identity features from non-identity features, ensuring the swapped face $\hat{x}$ and input template $x$ in the same domain as well as maintaining the high-resolution output of the original decoder.

\subsection{Monocular Facial Reflectance Inference}
\label{sec:3-3}
Once multi-domain codebooks and the identity-conditioned swapper are trained, we design an ID2Reflectance framework to reconstruct the reflectance maps as shown in Fig.~\ref{fig:pipeline}. Unlike previous facial reflectance prediction methods~\cite{Li20,Lattas20} designed in the UV space, our framework first synthesizes multi-view identity-conditioned reflectance images in the wrapped space and then stitches them together to obtain the final reflectance maps.

\noindent{\bf Multi-view Reflectance Swapping.}
Given an input face $\mathbf{I_{id}}$, we employ the identity similarity to search for the closet reflectance template (\eg, diffuse albedo $\mathbf{D}$, specular albedo $\mathbf{S}$, roughness $\mathbf{R}$, and surface normal $\mathbf{N}$). For each of these four reflectance domains, we select the three fixed views (\ie, left, frontal, and right). These 12 template reflectance images provide the condition of the facial attribute (\eg, domain and pose) for the quantized autoencoder, while the $\mathbf{I_{id}}$ provides the information of identity. As illustrated in Fig.~\ref{fig:pipeline}, the input face $\mathbf{I_{id}}$ is passed through the ArcFace embedding network $\mathcal{A}$ to extract the identity feature~${z_{id}\in\mathbb{R}^{\mathrm{1} \times 512}}$, while the 12 multi-view reflectance templates separately go through the encoder, multiple codebook quantization, latent representation fusion, and identity-conditioned generator to obtain multi-view identity-conditioned reflectance data.

\noindent{\bf Multi-view UV Stitching.}
To stitch the three-view reflectance images to get the reflectance map, we use Deep3D~\cite{Deng19} to estimate the shape in each view of the diffuse albedos, establish dense correspondences among three views, and unfold the facial reflectance in the UV space. Besides, we employ a face parsing model~\cite{lin2021roi} to predict the facial region, excluding non-facial areas (\eg, hats) for the unwrapped texture UV maps. To blend three-view diffuse albedos, we perform color matching in the YUV space~\cite{bai2023ffhquv} to merge the maps from the left and right views with the map from the frontal view. To obtain the specular albedo, roughness, and surface normal maps, we employ the same dense correspondences estimated for multi-view diffuse albedos. In this way, we obtain high-quality and identity-preserved facial reflectance assets as shown in Fig.~\ref{fig:pipeline}.

\section{Experiments}
\subsection{Implementation Details}
\label{Sec:4-1}
All our implementations are based on PyTorch~\cite{Paszke2019PyTorchAI} and Nvidia A6000 GPUs.
We employ the Adam~\cite{kingma2014adam} optimizer with a batch size of 16 for all training tasks in this paper.
For reflectance prior, we first train our models on FFHQ~\cite{karras2019style} and captured dataset, and all images are resized to $512\times512$ for training. We set the
latent code size as $16\times16$, and  
the codebook size as $1024\times256$. 
Our captured dataset contains 135 participants with gender, age, and race diversity. We randomly select 115 subjects for training and the rest 20 subjects for testing. 
In the first stage, we train the model on a mixture of FFHQ and reflectance data, with 
an initial learning rate of 8e-5 and a total iteration number of 700K.
In the second stage, we fine-tune each domain-specific codebook on the corresponding dataset and finally fine-tune the fusion module by a balanced sampling on FFHQ and captured data. For each fine-tuning step, the initial learning rate is set as 5e-5, and the iteration number is 100K.

To train the face swapper module, we choose VGGFace2HQ~\cite{cao2018vggface2} as our training set.
To improve the training data quality, we remove small images, improve the resolution by GFPGAN~\cite{wang2021gfpgan}, and resize images to $512\times512$ for training.
The pre-processed dataset contains 1.77 Million images from 8K identities.
Our swapper module $\mathcal{S}$ uses a fully connected architecture with random initialization. The target image is resized to $112\times112$~\cite{deng2020retinaface,guo2022sample,liu2022mogface,liu2023damofd} for ID feature embedding and the size of the final swapped reflectance is $512\times512$. 
Here, the initial learning rate is set as 7e-5, and the iteration number is set to 500K.

Our entire ID2Reflectance framework consists of identity embedding, multi-view reflectance swapping, dense correspondence estimation, and stitching. The total inference time is 3.8 seconds/face on the A6000 GPU. To visualize our relighting results, we employ the off-the-shelf geometry prediction network Deep3D~\cite{Deng19} to estimate the face shape. 

\subsection{Reflectance Reconstruction Results}
\label{Sec:4-2}

\begin{figure}[t]
    \centering
    \includegraphics[width=\linewidth]{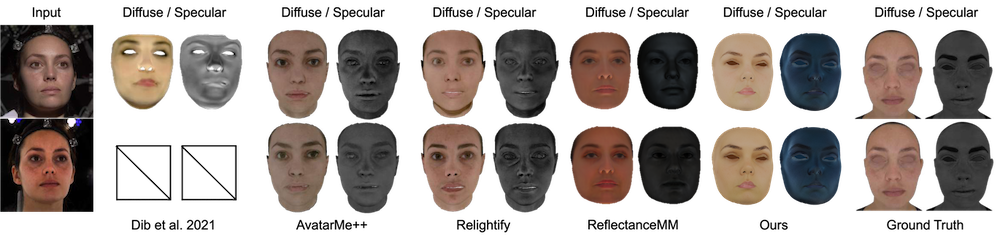}
    \caption{Comparison of diffuse and specular albedo reconstruction on Digital Emily project. From left to right: input image, Dib~\etal~\cite{dib2021towards}, AvatarMe++~\cite{lattas2021avatarme++}, Relightify~\cite{papantoniou2023relightify}, ReflectanceMM~\cite{han2023ReflectanceMM}, ours and ground-truth.}
    \label{fig:emily}
\end{figure}

\begin{figure}[t]
    \centering
    \includegraphics[width=\linewidth]{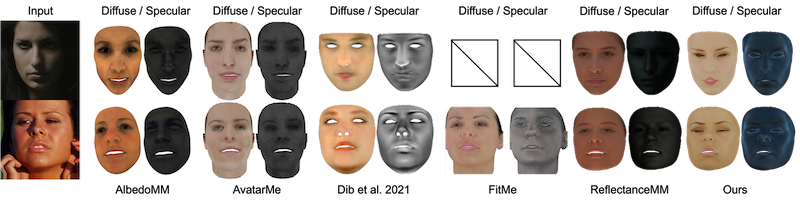}
    \caption{Comparison with recent single image reflectance prediction methods. From left to right: input image, AlbedoMM~\cite{smith2020AlbedoMM}, AvatarMe~\cite{Lattas20}, Dib~\etal~\cite{dib2021towards}, FitMe~\cite{lattas2023fitme}, ReflectanceMM~\cite{han2023ReflectanceMM} and ours.}
    \label{fig:in-the-wild-image}
\end{figure}

\begin{figure*}[t]
    \centering
    \includegraphics[width=\linewidth]{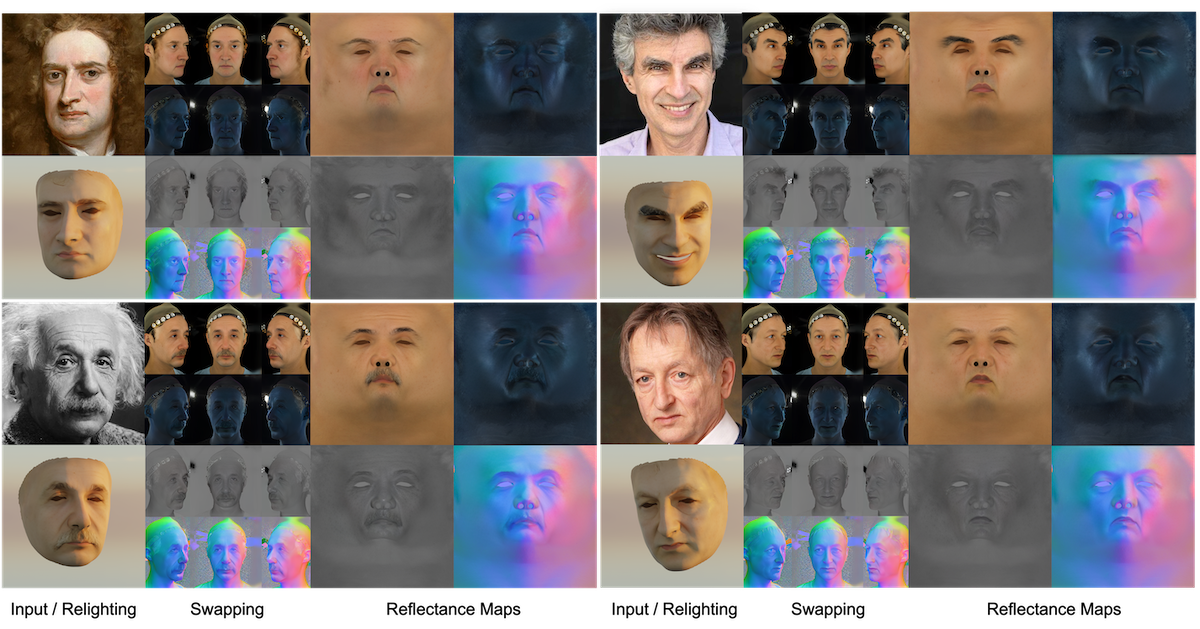}
    \caption{Qualitative results on unconstrained images. Given an input image, we initially obtain multi-domain three-view images through reflectance swapping and then stitch them into high-quality reflectance maps for relighting.}
    \label{fig:qual_workflow}
\end{figure*}

\begin{figure}[t]
    \centering
    \includegraphics[width=\linewidth]{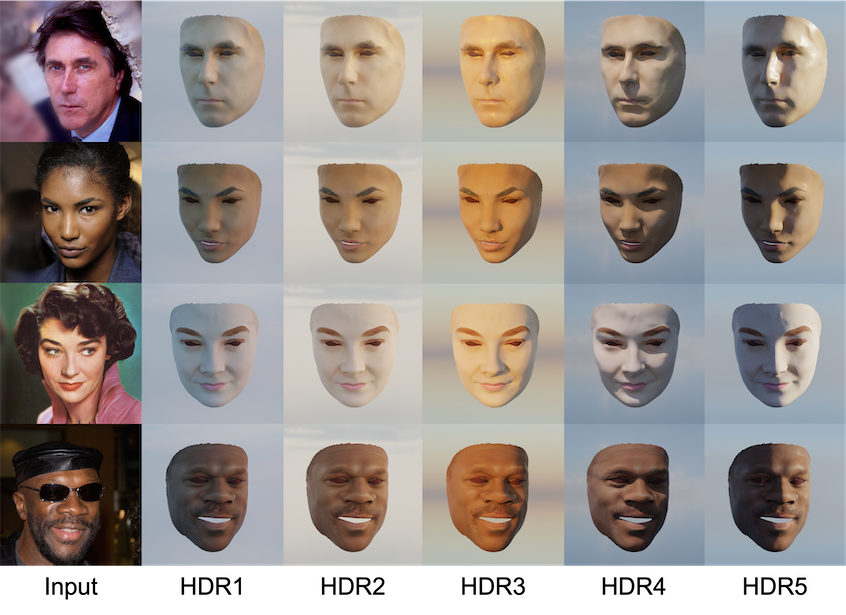}
    \caption{Relighting results for people from different races. Each column of images maintains the same HDR illumination.}
    \label{fig:relighting}
\end{figure}

We first evaluate our method in the task of monocular facial reflectance prediction.
Following recent state-of-the-art methods~\cite{dib2021towards,lattas2021avatarme++,lattas2023fitme,han2023ReflectanceMM,papantoniou2023relightify}, we perform visual comparisons on the well-known Digital Emily and in-the-wild data, respectively. In Fig.~\ref{fig:emily} and~\ref{fig:in-the-wild-image}, our method yields consistent and identity-preserved results regardless of the illumination variation. The minor lipstick artifacts in our results are due to the VGGFace2HQ dataset. This is because the restorer, GFPGAN, alters the makeup of some female subjects.

In our approach, we select the reflectance template with the highest similarity from the template library as the input of the encoder.
To evaluate the impact of template selection, we randomly select a Chinese male adult template and provide step-wise qualitative results in Fig.~\ref{fig:qual_workflow}.
Even though the reflectance template is fixed, our method still attains a high level of identity preservation. The only difference is the skin tone colour in diffuse albedo.
As shown in Fig.~\ref{fig:relighting}, we visualize some relighting results of our method. Given the ethnicity-diverse template library, our method is capable of reconstructing a wide range of races, producing relighting results at superior quality.

\subsection{Albedo Reconstruction Results}
\label{Sec:4-3}
\begin{figure}[t]
    \centering
    \includegraphics[width=\linewidth]{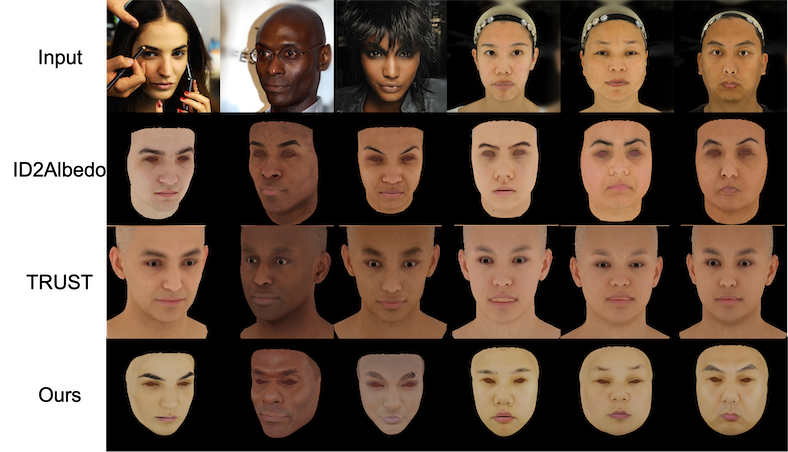}
    \caption{Comparisons of albedo estimation on in-the-wild and constrained images. Input images are from CelebAMask-HQ~\cite{CelebAMask-HQ} and our captured test set.}
    \label{fig:uvspace}
\end{figure}

\begin{figure}[t]
    \centering
    \includegraphics[width=\linewidth]{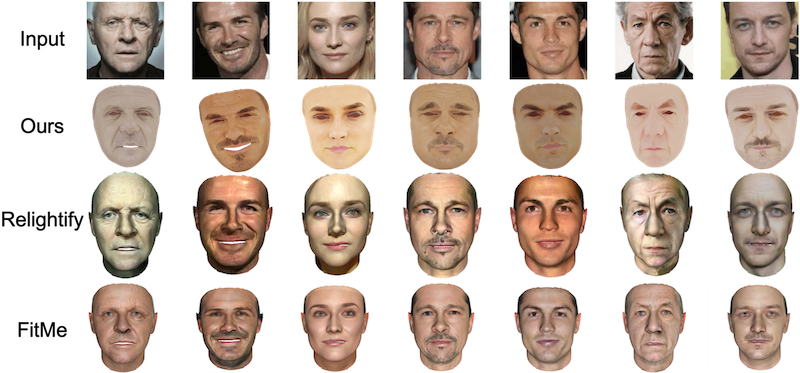}
    \caption{Comparisons of albedo estimation on in-the-wild and constrained images. From top to bottom: Input, Ours, Relightify~\cite{papantoniou2023relightify}, and FitMe~\cite{lattas2023fitme}. }
    \label{fig:in-the-wild}
\end{figure}

We employ CelebAMask-HQ~\cite{CelebAMask-HQ} and our test set (ground truth reflectance data captured from 20 participants) to evaluate our facial diffuse albedo prediction. As shown in Fig.~\ref{fig:uvspace} and~\ref{fig:in-the-wild}, we compare these results with recent state-of-the-art albedo estimation methods. 
Even though some of the input images contain occlusions, the albedos estimated by our method are robust under these occlusions, exhibiting realism, identity preservation, and race consistency.
In addition, we quantitatively compare our method with previous methods using identity similarity, PSNR, SSIM, and LPIPS~\cite{zhang2018perceptual} metrics.
To calculate identity similarity for TRUST~\cite{feng2022towards}, ID2Albedo~\cite{id2albedo}, and our method, we first overlay the diffuse albedo onto the original input image and then compute the cosine similarity between the overlaid albedo and the input face by using a pre-trained CosFace~\cite{wang2018cosface} model. 
Tab.~\ref{tab:albedo} demonstrates that our method achieves the best results in all image-level metrics.

\begin{table}[t]
\centering
\resizebox{0.48\textwidth}{!}{
\begin{tabular}{@{}l|cc|cc@{}}
\toprule
\multicolumn{1}{c|}{Methods} & PSNR$\uparrow$ & \multicolumn{1}{c}{SSIM$\uparrow$} & LPIPS$\downarrow$ & ID$\uparrow$~                \\ \midrule
~TRUST~\cite{feng2022towards} & 21.63 & 0.852 & 0.2014 &   0.478 ~\\
~ID2Albedo~\cite{id2albedo} & 23.72 & 0.884 & 0.1549 &   0.532 ~\\ \midrule
~Ours  & \textbf{28.47} & \textbf{0.923} & \textbf{0.1248} & \textbf{0.735} ~\\ \bottomrule
\end{tabular}}
\caption{Comparisons of our method with previous methods~\cite{id2albedo,feng2022towards} on albedo estimation.
Note that we compute PSNR and SSIM on the captured test set, while the ID similarity and LPIPS metrics are computed on the CelebAMask-HQ~\cite{CelebAMask-HQ} dataset.}
\label{tab:albedo}
\end{table}

\begin{table}[t]
\centering
\resizebox{0.48\textwidth}{!}{
\begin{tabular}{@{}l|c|c|c|c@{}}
\toprule
\multicolumn{1}{c|}{Configs} & Diffuse  & Specular  & Roughness & Normal ~ \\ \midrule
~Joint codebook         & 24.87 & 20.95 & 21.31  & 20.56  ~\\ 
~Multi-domain codebooks & 31.62 & 30.96 & 31.59  & 30.32 ~\\ \midrule
~Fixed Template   & 25.26 & 26.44 & 29.56  & 25.77  ~\\ 
~Closest Template   & 28.47 & 26.68 & 30.32 & 26.83  ~\\ \bottomrule
\end{tabular}}
\caption{Comparison of ID2Reflectance framework under different configurations. (1) joint codebook v.s. multi-domain codebooks for reflectance reconstruction, and (2) fixed swapping template v.s. closest swapping template for identity-conditioned reflectance prediction. We calculate the average PSNR between the reconstructed and the ground truth reflectance maps on our test set.}
\label{tab:ID2ReflectanceConfig}
\end{table}

\subsection{Ablation Studies}
\label{Sec:4-4}
\begin{figure}[t]
    \centering
    \includegraphics[width=\linewidth]{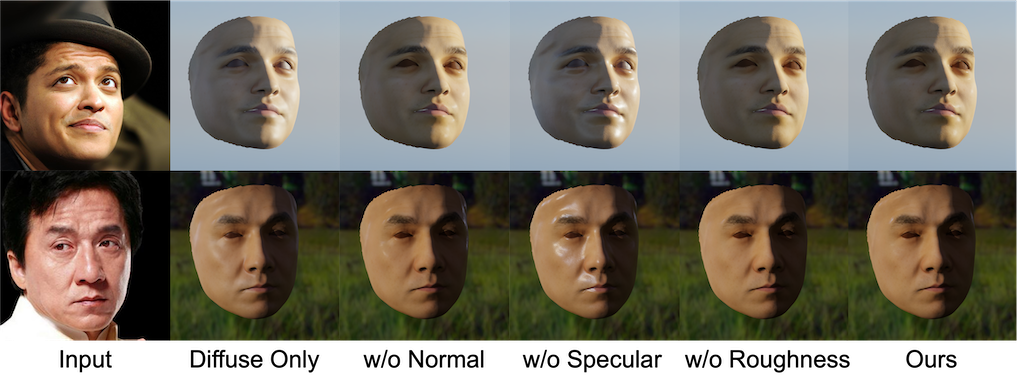}
    \caption{Ablation studies on each reflectance component predicted by our model.}
    \label{fig:ablation_pbr}
\end{figure}

\noindent{\bf ID2Reflectance Framework.}
In the first and second rows of Tab.~\ref{tab:ID2ReflectanceConfig}, we reconstruct the three-view reflectance images directly using the captured data and stitch them together to get complete UV maps. As we can see, the multi-domain codebook learning significantly outperforms the joint codebook learning. In the third and fourth rows, we compare reflectance swapping under the fixed template and the closest template by using multi-domain codebooks. The results indicate that the fixed template setting obviously underperforms the closest template setting on diffuse albedo estimation. As depicted in Fig.~\ref{fig:qual_workflow}, using a fixed template results in a skin tone gap in diffuse albedos when the target and source faces are from two different races. In addition, we perform ablation studies on each reflectance component predicted by our method in Fig.~\ref{fig:ablation_pbr}. The results verify that the predicted normal, specular, roughness maps are beneficial to improve the quality of physically based rendering.
While good results were achieved, it's important to note that our reconstructed PBR maps may not be physically plausible, as they weren't supervised using ground truth PBR maps.

\begin{figure}[t]
    \centering
    \includegraphics[width=\linewidth]{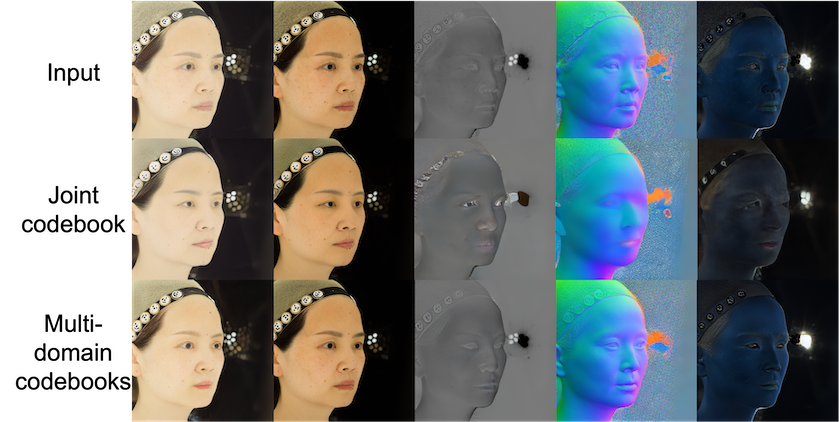}
    \caption{Reconstruction comparison of using joint and multi-domain codebooks. Inputs are the same faces from PBR domains.}
    \label{fig:codebook}
\end{figure}

\noindent{\bf Multi-domain Codebook.}
As shown in Fig.~\ref{fig:codebook}, the input and reconstruction results from the RGB and reflectance domains demonstrate that the joint codebook has limited capability in recovering cross-domain data, leading to visible artifacts and an over-smoothed face structure. By contrast, our multi-domain codebooks achieve finer facial structure and better reconstruction quality.

\begin{figure}[t]
    \centering
    \includegraphics[width=\linewidth]{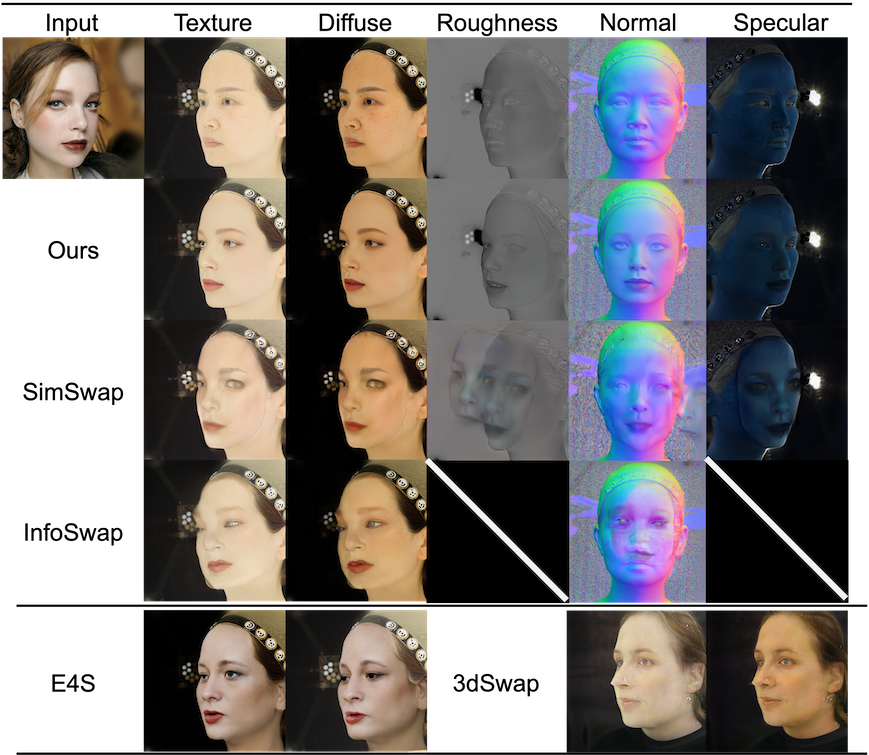}
    \caption{Cross-domain swapping comparison with other methods (\eg, SimSwap~\cite{chen2020simswap}, InfoSwap~\cite{gao2021inforswap}, E4S~\cite{liu2023fine}, and 3dSwap~\cite{li20233dfaceswapping}). Input RGB face provides the source identity while other reflectance data are the target. }
    \label{fig:SwapperRGB}
\end{figure}

\noindent{\bf Swapper Module.}
In Fig.~\ref{fig:SwapperRGB}, we compare our swapping module with state-of-the-art methods.
As we can see, SimSwap~\cite{chen2020simswap} and InfoSwap~\cite{gao2021inforswap} encounter great difficulties in the reflectance domain, generating obvious noises and artifacts in the results. Moreover, StyleGAN-based swapping methods (\eg, E4S~\cite{liu2023fine} and 3dSwap~\cite{li20233dfaceswapping}) can only run in the texture and diffuse domains as the patterns in StyleGAN~\cite{karras2019style} and EG3D~\cite{chan2022eg3d} are insufficient to cover other reflectance domains. By contrast, our approach achieves consistent and high-quality swapping results on both RGB and reflectance domains.

\section{Conclusions and Discussions}
In this paper, we present a novel monocular facial reflectance reconstruction method. By learning high-quality multi-domain discrete codebooks, we can obtain a reliable reflection prior from limited captured data. Our model utilizes identity features as conditions to reconstruct multi-view reflectance images directly from the multi-domain codebooks and then stitches them together into a complete reflectance map. Experiments demonstrate that our method exhibits good qualitative and quantitative performance, and excellent generalization to real-world images.

\noindent{\bf Broader Impacts.}
Although not the purpose of our work, monocular facial reflectance reconstruction may potentially be abused. Nevertheless, our method can be used for high-quality reconstruction and rendering, opening new avenues for faster avatar creation and promotion of the metaverse.

\paragraph{Acknowledgements} This work was supported in part by NSFC (62322113, 62376156, 62101325, 62201342), Shanghai Municipal Science and Technology Major Project (2021SHZDZX0102), and the Fundamental Research Funds for the Central Universities.
\clearpage
{
    \small
    \bibliographystyle{ieeenat_fullname}
    \bibliography{main}

\begin{thebibliography}{77}
\providecommand{\natexlab}[1]{#1}
\providecommand{\url}[1]{\texttt{#1}}
\expandafter\ifx\csname urlstyle\endcsname\relax
  \providecommand{\doi}[1]{doi: #1}\else
  \providecommand{\doi}{doi: \begingroup \urlstyle{rm}\Url}\fi

\bibitem[Bai et~al.(2023)Bai, Kang, Zhang, Pan, and Bao]{bai2023ffhquv}
Haoran Bai, Di Kang, Haoxian Zhang, Jinshan Pan, and Linchao Bao.
\newblock Ffhq-uv: Normalized facial uv-texture dataset for 3d face reconstruction.
\newblock In \emph{CVPR}, 2023.

\bibitem[Bai et~al.(2021)Bai, Cui, Liu, and Tan]{Bai21}
Ziqian Bai, Zhaopeng Cui, Xiaoming Liu, and Ping Tan.
\newblock Riggable {3D} face reconstruction via in-network optimization.
\newblock In \emph{CVPR}, 2021.

\bibitem[Bitouk et~al.(2008)Bitouk, Kumar, Dhillon, Belhumeur, and Nayar]{bitouk2008faceswapping}
Dmitri Bitouk, Neeraj Kumar, Samreen Dhillon, Peter Belhumeur, and Shree~K Nayar.
\newblock Face swapping: automatically replacing faces in photographs.
\newblock In \emph{SIGGRAPH}, 2008.

\bibitem[Blanz and Vetter(1999)]{Blanz99}
Volker Blanz and Thomas Vetter.
\newblock A morphable model for the synthesis of 3d faces.
\newblock In \emph{SIGGRAPH}, 1999.

\bibitem[Booth et~al.(2016)Booth, Roussos, Zafeiriou, Ponniah, and Dunaway]{booth20163d}
James Booth, Anastasios Roussos, Stefanos Zafeiriou, Allan Ponniah, and David Dunaway.
\newblock A 3d morphable model learnt from 10,000 faces.
\newblock In \emph{CVPR}, 2016.

\bibitem[Cao et~al.(2018)Cao, Shen, Xie, Parkhi, and Zisserman]{cao2018vggface2}
Qiong Cao, Li Shen, Weidi Xie, Omkar~M Parkhi, and Andrew Zisserman.
\newblock Vggface2: A dataset for recognising faces across pose and age.
\newblock In \emph{FG}, 2018.

\bibitem[Chan et~al.(2022)Chan, Lin, Chan, Nagano, Pan, De~Mello, Gallo, Guibas, Tremblay, Khamis, et~al.]{chan2022eg3d}
Eric~R Chan, Connor~Z Lin, Matthew~A Chan, Koki Nagano, Boxiao Pan, Shalini De~Mello, Orazio Gallo, Leonidas~J Guibas, Jonathan Tremblay, Sameh Khamis, et~al.
\newblock Efficient geometry-aware 3d generative adversarial networks.
\newblock In \emph{CVPR}, 2022.

\bibitem[Chen et~al.(2020)Chen, Chen, Ni, and Ge]{chen2020simswap}
Renwang Chen, Xuanhong Chen, Bingbing Ni, and Yanhao Ge.
\newblock Simswap: An efficient framework for high fidelity face swapping.
\newblock In \emph{ACM MM}, 2020.

\bibitem[Debevec et~al.(2000)Debevec, Hawkins, Tchou, Duiker, Sarokin, and Sagar]{debevec2000acquiring}
Paul Debevec, Tim Hawkins, Chris Tchou, Haarm-Pieter Duiker, Westley Sarokin, and Mark Sagar.
\newblock Acquiring the reflectance field of a human face.
\newblock In \emph{Computer Graphics and Interactive Techniques}, 2000.

\bibitem[Deng et~al.(2018)Deng, Cheng, Xue, Zhou, and Zafeiriou]{deng2018uv}
Jiankang Deng, Shiyang Cheng, Niannan Xue, Yuxiang Zhou, and Stefanos Zafeiriou.
\newblock Uv-gan: Adversarial facial uv map completion for pose-invariant face recognition.
\newblock In \emph{CVPR}, 2018.

\bibitem[Deng et~al.(2019{\natexlab{a}})Deng, Guo, Xue, and Zafeiriou]{deng2019arcface}
Jiankang Deng, Jia Guo, Niannan Xue, and Stefanos Zafeiriou.
\newblock Arcface: Additive angular margin loss for deep face recognition.
\newblock In \emph{CVPR}, 2019{\natexlab{a}}.

\bibitem[Deng et~al.(2020)Deng, Guo, Ververas, Kotsia, and Zafeiriou]{deng2020retinaface}
Jiankang Deng, Jia Guo, Evangelos Ververas, Irene Kotsia, and Stefanos Zafeiriou.
\newblock Retinaface: Single-shot multi-level face localisation in the wild.
\newblock In \emph{CVPR}, 2020.

\bibitem[Deng et~al.(2019{\natexlab{b}})Deng, Yang, Xu, Chen, Jia, and Tong]{Deng19}
Yu Deng, Jiaolong Yang, Sicheng Xu, Dong Chen, Yunde Jia, and Xin Tong.
\newblock Accurate {3D} face reconstruction with weakly-supervised learning: From single image to image set.
\newblock In \emph{CVPRW}, 2019{\natexlab{b}}.

\bibitem[Dib et~al.(2021)Dib, Thebault, Ahn, Gosselin, Theobalt, and Chevallier]{dib2021towards}
Abdallah Dib, Cedric Thebault, Junghyun Ahn, Philippe-Henri Gosselin, Christian Theobalt, and Louis Chevallier.
\newblock Towards high fidelity monocular face reconstruction with rich reflectance using self-supervised learning and ray tracing.
\newblock In \emph{ICCV}, 2021.

\bibitem[Egger et~al.(2020)Egger, Smith, Tewari, Wuhrer, Zollh{\"{o}}fer, Beeler, Bernard, Bolkart, Kortylewski, Romdhani, Theobalt, Blanz, and Vetter]{Egger2020}
Bernhard Egger, William A.~P. Smith, Ayush Tewari, Stefanie Wuhrer, Michael Zollh{\"{o}}fer, Thabo Beeler, Florian Bernard, Timo Bolkart, Adam Kortylewski, Sami Romdhani, Christian Theobalt, Volker Blanz, and Thomas Vetter.
\newblock {3D} morphable face models - past, present, and future.
\newblock \emph{TOG}, 2020.

\bibitem[Esser et~al.(2021)Esser, Rombach, and Ommer]{esser2021taming}
Patrick Esser, Robin Rombach, and Bjorn Ommer.
\newblock Taming transformers for high-resolution image synthesis.
\newblock In \emph{CVPR}, 2021.

\bibitem[Feng et~al.(2022)Feng, Bolkart, Tesch, Black, and Abrevaya]{feng2022towards}
Haiwen Feng, Timo Bolkart, Joachim Tesch, Michael~J Black, and Victoria Abrevaya.
\newblock Towards racially unbiased skin tone estimation via scene disambiguation.
\newblock In \emph{ECCV}, 2022.

\bibitem[Feng et~al.(2021)Feng, Feng, Black, and Bolkart]{Feng2021}
Yao Feng, Haiwen Feng, Michael~J Black, and Timo Bolkart.
\newblock Learning an animatable detailed 3d face model from in-the-wild images.
\newblock \emph{TOG}, 2021.

\bibitem[Gao et~al.(2021)Gao, Huang, Fu, Li, and He]{gao2021inforswap}
Gege Gao, Huaibo Huang, Chaoyou Fu, Zhaoyang Li, and Ran He.
\newblock Information bottleneck disentanglement for identity swapping.
\newblock In \emph{CVPR}, 2021.

\bibitem[Gecer et~al.(2019)Gecer, Ploumpis, Kotsia, and Zafeiriou]{Gecer19:ganfit}
Baris Gecer, Stylianos Ploumpis, Irene Kotsia, and Stefanos Zafeiriou.
\newblock Ganfit: Generative adversarial network fitting for high fidelity 3d face reconstruction.
\newblock In \emph{CVPR}, 2019.

\bibitem[Gecer et~al.(2020)Gecer, Lattas, Ploumpis, Deng, Papaioannou, Moschoglou, and Zafeiriou]{gecer2020synthesizing}
Baris Gecer, Alexandros Lattas, Stylianos Ploumpis, Jiankang Deng, Athanasios Papaioannou, Stylianos Moschoglou, and Stefanos Zafeiriou.
\newblock Synthesizing coupled 3d face modalities by trunk-branch generative adversarial networks.
\newblock In \emph{ECCV}, 2020.

\bibitem[Ghosh et~al.(2011)Ghosh, Fyffe, Tunwattanapong, Busch, Yu, and Debevec]{ghosh2011multiview}
Abhijeet Ghosh, Graham Fyffe, Borom Tunwattanapong, Jay Busch, Xueming Yu, and Paul Debevec.
\newblock Multiview face capture using polarized spherical gradient illumination.
\newblock \emph{TOG}, 2011.

\bibitem[Goodfellow et~al.(2014)Goodfellow, Pouget-Abadie, Mirza, Xu, Warde-Farley, Ozair, Courville, and Bengio]{goodfellow2014generative}
Ian Goodfellow, Jean Pouget-Abadie, Mehdi Mirza, Bing Xu, David Warde-Farley, Sherjil Ozair, Aaron Courville, and Yoshua Bengio.
\newblock Generative adversarial nets.
\newblock In \emph{NeurIPS}, 2014.

\bibitem[Guo et~al.(2022)Guo, Deng, Lattas, and Zafeiriou]{guo2022sample}
Jia Guo, Jiankang Deng, Alexandros Lattas, and Stefanos Zafeiriou.
\newblock Sample and computation redistribution for efficient face detection.
\newblock In \emph{ICLR}, 2022.

\bibitem[Han et~al.(2023{\natexlab{a}})Han, Wang, and Xu]{han2023ReflectanceMM}
Yuxuan Han, Zhibo Wang, and Feng Xu.
\newblock Learning a 3d morphable face reflectance model from low-cost data.
\newblock In \emph{CVPR}, 2023{\natexlab{a}}.

\bibitem[Han et~al.(2023{\natexlab{b}})Han, Wang, and Xu]{han2023learning}
Yuxuan Han, Zhibo Wang, and Feng Xu.
\newblock Learning a 3d morphable face reflectance model from low-cost data.
\newblock In \emph{CVPR}, 2023{\natexlab{b}}.

\bibitem[Huang and Belongie(2017)]{huang2017arbitrary}
Xun Huang and Serge Belongie.
\newblock Arbitrary style transfer in real-time with adaptive instance normalization.
\newblock In \emph{ICCV}, 2017.

\bibitem[Isola et~al.(2017)Isola, Zhu, Zhou, and Efros]{isola2017image}
Phillip Isola, Jun-Yan Zhu, Tinghui Zhou, and Alexei~A Efros.
\newblock Image-to-image translation with conditional adversarial networks.
\newblock In \emph{CVPR}, 2017.

\bibitem[Karras et~al.(2018)Karras, Aila, Laine, and Lehtinen]{karras2017progressive}
Tero Karras, Timo Aila, Samuli Laine, and Jaakko Lehtinen.
\newblock Progressive growing of gans for improved quality, stability, and variation.
\newblock In \emph{ICLR}, 2018.

\bibitem[Karras et~al.(2019)Karras, Laine, and Aila]{karras2019style}
Tero Karras, Samuli Laine, and Timo Aila.
\newblock A style-based generator architecture for generative adversarial networks.
\newblock In \emph{CVPR}, 2019.

\bibitem[Karras et~al.(2020)Karras, Laine, Aittala, Hellsten, Lehtinen, and Aila]{Karras2019stylegan2}
Tero Karras, Samuli Laine, Miika Aittala, Janne Hellsten, Jaakko Lehtinen, and Timo Aila.
\newblock Analyzing and improving the image quality of stylegan.
\newblock In \emph{CVPR}, 2020.

\bibitem[Kingma and Ba(2015)]{kingma2014adam}
Diederik~P Kingma and Jimmy Ba.
\newblock Adam: A method for stochastic optimization.
\newblock In \emph{ICLR}, 2015.

\bibitem[Lattas et~al.(2020)Lattas, Moschoglou, Gecer, Ploumpis, Triantafyllou, Ghosh, and Zafeiriou]{Lattas20}
Alexandros Lattas, Stylianos Moschoglou, Baris Gecer, Stylianos Ploumpis, Vasileios Triantafyllou, Abhijeet Ghosh, and Stefanos Zafeiriou.
\newblock {AvatarMe}: Realistically renderable {3D} facial reconstruction.
\newblock In \emph{CVPR}, 2020.

\bibitem[Lattas et~al.(2021)Lattas, Moschoglou, Ploumpis, Gecer, Ghosh, and Zafeiriou]{lattas2021avatarme++}
Alexandros Lattas, Stylianos Moschoglou, Stylianos Ploumpis, Baris Gecer, Abhijeet Ghosh, and Stefanos~P Zafeiriou.
\newblock Avatarme$^{++}$: Facial shape and {BRDF} inference with photorealistic rendering-aware {GANs}.
\newblock \emph{TPAMI}, 2021.

\bibitem[Lattas et~al.(2023)Lattas, Moschoglou, Ploumpis, Gecer, Deng, and Zafeiriou]{lattas2023fitme}
Alexandros Lattas, Stylianos Moschoglou, Stylianos Ploumpis, Baris Gecer, Jiankang Deng, and Stefanos Zafeiriou.
\newblock Fitme: Deep photorealistic 3d morphable model avatars.
\newblock In \emph{CVPR}, 2023.

\bibitem[Lee et~al.(2020)Lee, Liu, Wu, and Luo]{CelebAMask-HQ}
Cheng-Han Lee, Ziwei Liu, Lingyun Wu, and Ping Luo.
\newblock Maskgan: Towards diverse and interactive facial image manipulation.
\newblock In \emph{CVPR}, 2020.

\bibitem[Lee and Lee(2020)]{lee2020uncertainty}
Gun-Hee Lee and Seong-Whan Lee.
\newblock Uncertainty-aware mesh decoder for high fidelity 3d face reconstruction.
\newblock In \emph{CVPR}, 2020.

\bibitem[Li et~al.(2020{\natexlab{a}})Li, Bao, Yang, Chen, and Wen]{li2020faceshifter}
Lingzhi Li, Jianmin Bao, Hao Yang, Dong Chen, and Fang Wen.
\newblock Faceshifter: Towards high fidelity and occlusion aware face swapping.
\newblock In \emph{CVPR}, 2020{\natexlab{a}}.

\bibitem[Li et~al.(2020{\natexlab{b}})Li, Bladin, Zhao, Chinara, Ingraham, Xiang, Ren, Prasad, Kishore, Xing, and Li]{li2020learning}
Ruilong Li, Karl Bladin, Yajie Zhao, Chinmay Chinara, Owen Ingraham, Pengda Xiang, Xinglei Ren, Pratusha Prasad, Bipin Kishore, Jun Xing, and Hao Li.
\newblock Learning formation of physically-based face attributes.
\newblock In \emph{CVPR}, 2020{\natexlab{b}}.

\bibitem[Li et~al.(2020{\natexlab{c}})Li, Bladin, Zhao, Chinara, Ingraham, Xiang, Ren, Prasad, Kishore, Xing, et~al.]{Li20}
Ruilong Li, Karl Bladin, Yajie Zhao, Chinmay Chinara, Owen Ingraham, Pengda Xiang, Xinglei Ren, Pratusha Prasad, Bipin Kishore, Jun Xing, et~al.
\newblock Learning formation of physically-based face attributes.
\newblock In \emph{CVPR}, 2020{\natexlab{c}}.

\bibitem[Li et~al.(2023)Li, Ma, Yan, Zhu, and Yang]{li20233dfaceswapping}
Yixuan Li, Chao Ma, Yichao Yan, Wenhan Zhu, and Xiaokang Yang.
\newblock 3d-aware face swapping.
\newblock In \emph{CVPR}, 2023.

\bibitem[Lin et~al.(2021)Lin, Shen, Wang, and Pantic]{lin2021roi}
Yiming Lin, Jie Shen, Yujiang Wang, and Maja Pantic.
\newblock Roi tanh-polar transformer network for face parsing in the wild.
\newblock \emph{IVC}, 2021.

\bibitem[Liu et~al.(2023{\natexlab{a}})Liu, Jiang, Choi, and Gu]{liu2023learning}
Kechun Liu, Yitong Jiang, Inchang Choi, and Jinwei Gu.
\newblock Learning image-adaptive codebooks for class-agnostic image restoration.
\newblock In \emph{ICCV}, 2023{\natexlab{a}}.

\bibitem[Liu et~al.(2022)Liu, Wang, Deng, Zhou, Sun, and Li]{liu2022mogface}
Yang Liu, Fei Wang, Jiankang Deng, Zhipeng Zhou, Baigui Sun, and Hao Li.
\newblock Mogface: Towards a deeper appreciation on face detection.
\newblock In \emph{CVPR}, 2022.

\bibitem[Liu et~al.(2023{\natexlab{b}})Liu, Deng, Wang, Shang, Xie, and Sun]{liu2023damofd}
Yang Liu, Jiankang Deng, Fei Wang, Lei Shang, Xuansong Xie, and Baigui Sun.
\newblock Damofd: Digging into backbone design on face detection.
\newblock In \emph{ICLR}, 2023{\natexlab{b}}.

\bibitem[Liu et~al.(2021)Liu, Lin, Cao, Hu, Wei, Zhang, Lin, and Guo]{liu2021swin}
Ze Liu, Yutong Lin, Yue Cao, Han Hu, Yixuan Wei, Zheng Zhang, Stephen Lin, and Baining Guo.
\newblock Swin transformer: Hierarchical vision transformer using shifted windows.
\newblock In \emph{ICCV}, 2021.

\bibitem[Liu et~al.(2023{\natexlab{c}})Liu, Li, Zhang, Wang, Zhang, Wang, and Nie]{liu2023fine}
Zhian Liu, Maomao Li, Yong Zhang, Cairong Wang, Qi Zhang, Jue Wang, and Yongwei Nie.
\newblock Fine-grained face swapping via regional gan inversion.
\newblock In \emph{CVPR}, 2023{\natexlab{c}}.

\bibitem[Luo et~al.(2021)Luo, Nagano, Kung, Xu, Wang, Wei, Hu, and Li]{luo2021normalized}
Huiwen Luo, Koki Nagano, Han-Wei Kung, Qingguo Xu, Zejian Wang, Lingyu Wei, Liwen Hu, and Hao Li.
\newblock Normalized avatar synthesis using stylegan and perceptual refinement.
\newblock In \emph{CVPR}, 2021.

\bibitem[Luo et~al.(2022)Luo, Zhu, He, Chu, Tai, Wang, and Yan]{luo2022styleface}
Yuchen Luo, Junwei Zhu, Keke He, Wenqing Chu, Ying Tai, Chengjie Wang, and Junchi Yan.
\newblock Styleface: Towards identity-disentangled face generation on megapixels.
\newblock In \emph{ECCV}, 2022.

\bibitem[Ma et~al.(2021)Ma, Simon, Saragih, Wang, Li, De~La~Torre, and Sheikh]{ma2021pixel}
Shugao Ma, Tomas Simon, Jason Saragih, Dawei Wang, Yuecheng Li, Fernando De~La~Torre, and Yaser Sheikh.
\newblock Pixel codec avatars.
\newblock In \emph{CVPR}, 2021.

\bibitem[Papantoniou et~al.(2023)Papantoniou, Lattas, Moschoglou, and Zafeiriou]{papantoniou2023relightify}
Foivos~Paraperas Papantoniou, Alexandros Lattas, Stylianos Moschoglou, and Stefanos Zafeiriou.
\newblock Relightify: Relightable 3d faces from a single image via diffusion models.
\newblock In \emph{ICCV}, 2023.

\bibitem[Paszke et~al.(2019)Paszke, Gross, Massa, Lerer, Bradbury, Chanan, Killeen, Lin, Gimelshein, Antiga, Desmaison, K{\"o}pf, Yang, DeVito, Raison, Tejani, Chilamkurthy, Steiner, Fang, Bai, and Chintala]{Paszke2019PyTorchAI}
Adam Paszke, Sam Gross, Francisco Massa, Adam Lerer, James Bradbury, Gregory Chanan, Trevor Killeen, Zeming Lin, Natalia Gimelshein, Luca Antiga, Alban Desmaison, Andreas K{\"o}pf, Edward Yang, Zach DeVito, Martin Raison, Alykhan Tejani, Sasank Chilamkurthy, Benoit Steiner, Lu Fang, Junjie Bai, and Soumith Chintala.
\newblock Pytorch: An imperative style, high-performance deep learning library.
\newblock In \emph{NeurIPS}, 2019.

\bibitem[Ren et~al.(2023{\natexlab{a}})Ren, Chen, Yao, Shum, and Wang]{ren2023reinforced}
Xiaohang Ren, Xingyu Chen, Pengfei Yao, Heung-Yeung Shum, and Baoyuan Wang.
\newblock Reinforced disentanglement for face swapping without skip connection.
\newblock In \emph{CVPR}, 2023{\natexlab{a}}.

\bibitem[Ren et~al.(2023{\natexlab{b}})Ren, Deng, Ma, Yan, and Yang]{id2albedo}
Xingyu Ren, Jiankang Deng, Chao Ma, Yichao Yan, and Xiaokang Yang.
\newblock Improving fairness in facial albedo estimation via visual-textual cues.
\newblock In \emph{CVPR}, 2023{\natexlab{b}}.

\bibitem[Rombach et~al.(2022)Rombach, Blattmann, Lorenz, Esser, and Ommer]{rombach2022stablediffusion}
Robin Rombach, Andreas Blattmann, Dominik Lorenz, Patrick Esser, and Björn Ommer.
\newblock High-resolution image synthesis with latent diffusion models.
\newblock In \emph{CVPR}, 2022.

\bibitem[Sauer et~al.(2021)Sauer, Chitta, M{\"{u}}ller, and Geiger]{Sauer2021ProjectedGAN}
Axel Sauer, Kashyap Chitta, Jens M{\"{u}}ller, and Andreas Geiger.
\newblock Projected gans converge faster.
\newblock In \emph{NeurIPS}, 2021.

\bibitem[Shang et~al.(2020)Shang, Shen, Li, Zhou, Zhen, Fang, and Quan]{Shang20}
Jiaxiang Shang, Tianwei Shen, Shiwei Li, Lei Zhou, Mingmin Zhen, Tian Fang, and Long Quan.
\newblock Self-supervised monocular {3D} face reconstruction by occlusion-aware multi-view geometry consistency.
\newblock In \emph{ECCV}, 2020.

\bibitem[Shiohara et~al.(2023)Shiohara, Yang, and Taketomi]{shiohara2023blendface}
Kaede Shiohara, Xingchao Yang, and Takafumi Taketomi.
\newblock Blendface: Re-designing identity encoders for face-swapping.
\newblock In \emph{CVPR}, 2023.

\bibitem[Simonyan and Zisserman(2015)]{simonyan2014very}
Karen Simonyan and Andrew Zisserman.
\newblock Very deep convolutional networks for large-scale image recognition.
\newblock In \emph{ICLR}, 2015.

\bibitem[Smith et~al.(2020)Smith, Seck, Dee, Tiddeman, Tenenbaum, and Egger]{smith2020AlbedoMM}
William A.~P. Smith, Alassane Seck, Hannah Dee, Bernard Tiddeman, Joshua Tenenbaum, and Bernhard Egger.
\newblock A morphable face albedo model.
\newblock In \emph{CVPR}, 2020.

\bibitem[Tewari et~al.(2019)Tewari, Bernard, Garrido, Bharaj, Elgharib, Seidel, P{\'e}rez, Zollhofer, and Theobalt]{Tewari19}
Ayush Tewari, Florian Bernard, Pablo Garrido, Gaurav Bharaj, Mohamed Elgharib, Hans-Peter Seidel, Patrick P{\'e}rez, Michael Zollhofer, and Christian Theobalt.
\newblock Fml: Face model learning from videos.
\newblock In \emph{CVPR}, 2019.

\bibitem[Tran and Liu(2018)]{tran2018nonlinear}
Luan Tran and Xiaoming Liu.
\newblock Nonlinear 3d face morphable model.
\newblock In \emph{CVPR}, 2018.

\bibitem[Van Den~Oord et~al.(2017)Van Den~Oord, Vinyals, et~al.]{van2017vqvae}
Aaron Van Den~Oord, Oriol Vinyals, et~al.
\newblock Neural discrete representation learning.
\newblock In \emph{NeurIPS}, 2017.

\bibitem[Vetter and Blanz(1998)]{vetter19983dmm}
Thomas Vetter and Volker Blanz.
\newblock Estimating coloured 3d face models from single images: An example based approach.
\newblock In \emph{ECCV}, 1998.

\bibitem[Wang et~al.(2018)Wang, Wang, Zhou, Ji, Gong, Zhou, Li, and Liu]{wang2018cosface}
Hao Wang, Yitong Wang, Zheng Zhou, Xing Ji, Dihong Gong, Jingchao Zhou, Zhifeng Li, and Wei Liu.
\newblock Cosface: Large margin cosine loss for deep face recognition.
\newblock In \emph{CVPR}, 2018.

\bibitem[Wang et~al.(2021{\natexlab{a}})Wang, Li, Zhang, and Shan]{wang2021gfpgan}
Xintao Wang, Yu Li, Honglun Zhang, and Ying Shan.
\newblock Towards real-world blind face restoration with generative facial prior.
\newblock In \emph{CVPR}, 2021{\natexlab{a}}.

\bibitem[Wang et~al.(2021{\natexlab{b}})Wang, Chen, Zhu, Chu, Tai, Wang, Li, Wu, Huang, and Ji]{wang2021hififace}
Yuhan Wang, Xu Chen, Junwei Zhu, Wenqing Chu, Ying Tai, Chengjie Wang, Jilin Li, Yongjian Wu, Feiyue Huang, and Rongrong Ji.
\newblock Hififace: 3d shape and semantic prior guided high fidelity face swapping.
\newblock In \emph{CVPR}, 2021{\natexlab{b}}.

\bibitem[Wen et~al.(2021)Wen, Liu, Raj, and Singh]{Wen21}
Yandong Wen, Weiyang Liu, Bhiksha Raj, and Rita Singh.
\newblock Self-supervised {3D} face reconstruction via conditional estimation.
\newblock In \emph{ICCV}, 2021.

\bibitem[Xu et~al.(2022{\natexlab{a}})Xu, Zhang, Han, Tian, Zeng, Tai, Wang, Wang, and Liu]{xu2022uniface}
Chao Xu, Jiangning Zhang, Yue Han, Guanzhong Tian, Xianfang Zeng, Ying Tai, Yabiao Wang, Chengjie Wang, and Yong Liu.
\newblock Designing one unified framework for high-fidelity face reenactment and swapping.
\newblock In \emph{ECCV}, 2022{\natexlab{a}}.

\bibitem[Xu et~al.(2022{\natexlab{b}})Xu, Zhang, Hua, He, Yi, and Liu]{xu2022regionfs}
Chao Xu, Jiangning Zhang, Miao Hua, Qian He, Zili Yi, and Yong Liu.
\newblock Region-aware face swapping.
\newblock In \emph{CVPR}, 2022{\natexlab{b}}.

\bibitem[Xu et~al.(2022{\natexlab{c}})Xu, Zhou, Hong, Liu, Liu, Guo, Han, Liu, Ding, and Wang]{xu2022styleswap}
Zhiliang Xu, Hang Zhou, Zhibin Hong, Ziwei Liu, Jiaming Liu, Zhizhi Guo, Junyu Han, Jingtuo Liu, Errui Ding, and Jingdong Wang.
\newblock Styleswap: Style-based generator empowers robust face swapping.
\newblock In \emph{ECCV}, 2022{\natexlab{c}}.

\bibitem[Yamaguchi et~al.(2018)Yamaguchi, Saito, Nagano, Zhao, Chen, Olszewski, Morishima, and Li]{Yamaguchi18}
Shugo Yamaguchi, Shunsuke Saito, Koki Nagano, Yajie Zhao, Weikai Chen, Kyle Olszewski, Shigeo Morishima, and Hao Li.
\newblock High-fidelity facial reflectance and geometry inference from an unconstrained image.
\newblock \emph{TOG}, 2018.

\bibitem[Zhang et~al.(2023)Zhang, Qiu, Lin, Zhang, Shi, Yang, Shi, Yang, Xu, and Yu]{zhang2023dreamface}
Longwen Zhang, Qiwei Qiu, Hongyang Lin, Qixuan Zhang, Cheng Shi, Wei Yang, Ye Shi, Sibei Yang, Lan Xu, and Jingyi Yu.
\newblock Dreamface: Progressive generation of animatable 3d faces under text guidance.
\newblock In \emph{SIGGRAPH}, 2023.

\bibitem[Zhang et~al.(2018)Zhang, Isola, Efros, Shechtman, and Wang]{zhang2018perceptual}
Richard Zhang, Phillip Isola, Alexei~A Efros, Eli Shechtman, and Oliver Wang.
\newblock The unreasonable effectiveness of deep features as a perceptual metric.
\newblock In \emph{CVPR}, 2018.

\bibitem[Zhou et~al.(2022)Zhou, Chan, Li, and Loy]{zhou2022towards}
Shangchen Zhou, Kelvin Chan, Chongyi Li, and Chen~Change Loy.
\newblock Towards robust blind face restoration with codebook lookup transformer.
\newblock In \emph{NeurIPS}, 2022.

\bibitem[Zhu et~al.(2021{\natexlab{a}})Zhu, Li, Wang, Xu, and Sun]{zhu2021megafs}
Yuhao Zhu, Qi Li, Jian Wang, Cheng-Zhong Xu, and Zhenan Sun.
\newblock One shot face swapping on megapixels.
\newblock In \emph{CVPR}, 2021{\natexlab{a}}.

\bibitem[Zhu et~al.(2021{\natexlab{b}})Zhu, Huang, Deng, Ye, Huang, Chen, Zhu, Yang, Lu, Du, et~al.]{zhu2021webface260m}
Zheng Zhu, Guan Huang, Jiankang Deng, Yun Ye, Junjie Huang, Xinze Chen, Jiagang Zhu, Tian Yang, Jiwen Lu, Dalong Du, et~al.
\newblock Webface260m: A benchmark unveiling the power of million-scale deep face recognition.
\newblock In \emph{CVPR}, 2021{\natexlab{b}}.

\end{thebibliography}
}

\maketitlesupplementary
\appendix
\setcounter{page}{1}
In this supplementary material, we first provide a detailed description of our light stage capture system (Sec.~\ref{ls}) and the datasets (Sec.~\ref{dataset}) used in the manuscript. Next, we offer an in-depth analysis of each module in the framework (\eg, codebook (Sec.~\ref{codebook}), swapper (Sec.~\ref{swapper}), stitching (Sec.~\ref{stitching})). Finally, we present more qualitative and quantitative comparisons (Sec.~\ref{vis}) to show our framework's fairness and robustness. 

\section{Light Stage Capture System}\label{ls}
\begin{figure}[t]
    \centering
    \includegraphics[width=\linewidth]{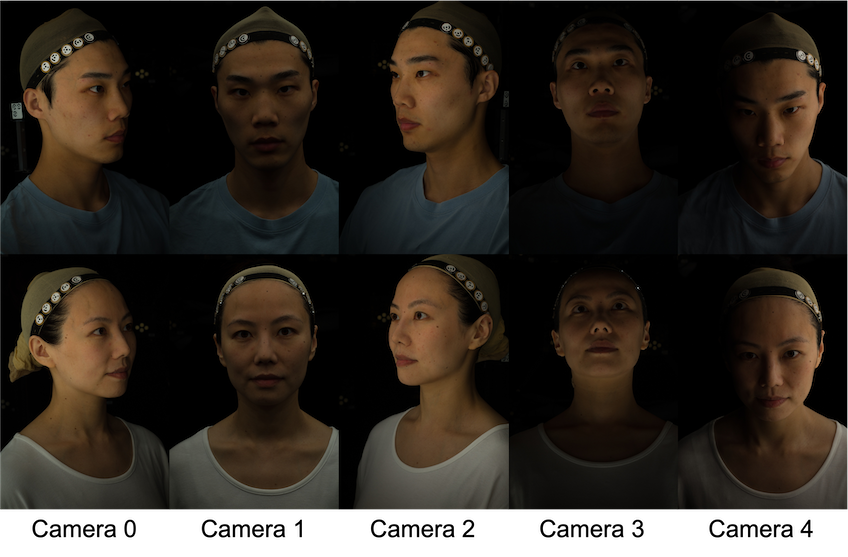}
    \caption{Raw data plots from 5 DSLR cameras with 25 MP resolution under different polarized light.}
    \label{fig:supp-pbr}
\end{figure}

Our Light Stage system, based on photometric stereo, achieves high-precision geometric reconstruction and pore-level reflectance map asset reconstruction.
Our system is composed of 46 DSLR cameras with 12MP resolution and 5 DSLR cameras with 25MP resolution. During capturing, the system performs 15 types of lighting in 2 seconds, including one full-light scenario where all cameras capture images for high-precision geometric reconstruction; the other 14 types of polarized light are captured only by the 5 cameras with 25MP resolution, used for computing pore-level diffuse, tangent normal, specular, and roughness. The original capture under different polarized light is shown in Fig.~\ref{fig:supp-pbr}.
In the processing phase, we first employed multi-view imaging for MVS modeling and engaged artists to manually register it to the BFM model for film-grade fitting accuracy. Then, the 8K assets obtained through photometric methods were re-topologized to the BFM topology, fitting the facial samples. The same manual process was applied to different expression results. To achieve accurate and consistent topology across different expressions, we marked specific points on areas like the eyes and corners of the mouth during capturing, which were manually removed by artists in post-processing.

\section{Public Datasets}\label{dataset}
We use the FFHQ~\cite{karras2019style} and CelebAMaskHQ~\cite{CelebAMask-HQ} datasets in addition to the captured dataset in Sec.~\ref{ls}.

\noindent{\bf FFHQ dataset.}
The FFHQ~\cite{karras2019style} dataset contains 70,000 high-quality facial images with a resolution of $1024\times1024$. It also displays diverse ages, ethnicities, and image backgrounds.

\noindent{\bf VGGFace2 dataset.}
The VGGFace2~\cite{cao2018vggface2} dataset is a large-scale face recognition dataset developed by the Visual Geometry Group at the University of Oxford. It contains over 3.3 million face images of over 9000 subjects, with an average of 362 images per subject. The images in VGGFace2 are highly varied in terms of age, ethnicity, and pose, providing a comprehensive resource for training and evaluating face recognition algorithms. Each subject in the dataset is captured in a wide range of conditions, including different lighting, expressions, and backgrounds.

\noindent{\bf CelebAMaskHQ dataset.}
CelebAMask-HQ~\cite{CelebAMask-HQ} Dataset is a high-quality facial image dataset designed for facial attribute analysis and face editing. It contains 30,000 high-resolution ($1024\times1024$) facial images selected from the original CelebA dataset.

\section{Details of Multi-domain Codebook}\label{codebook}
Our encoder $\mathcal{E}$ and decoder $\mathcal{G}$ consist of 12 residual blocks and 5 resizing layers, respectively, with a compression ratio of $r=2^5=32$. Besides, we set the number of items in the codebook N to 1024 and the dimensionality of each codebook d to 256. These settings are consistent with previous work, balancing the anti-degradation properties and computational costs in global modeling.

\begin{figure}[t]
    \centering
    \includegraphics[width=\linewidth]{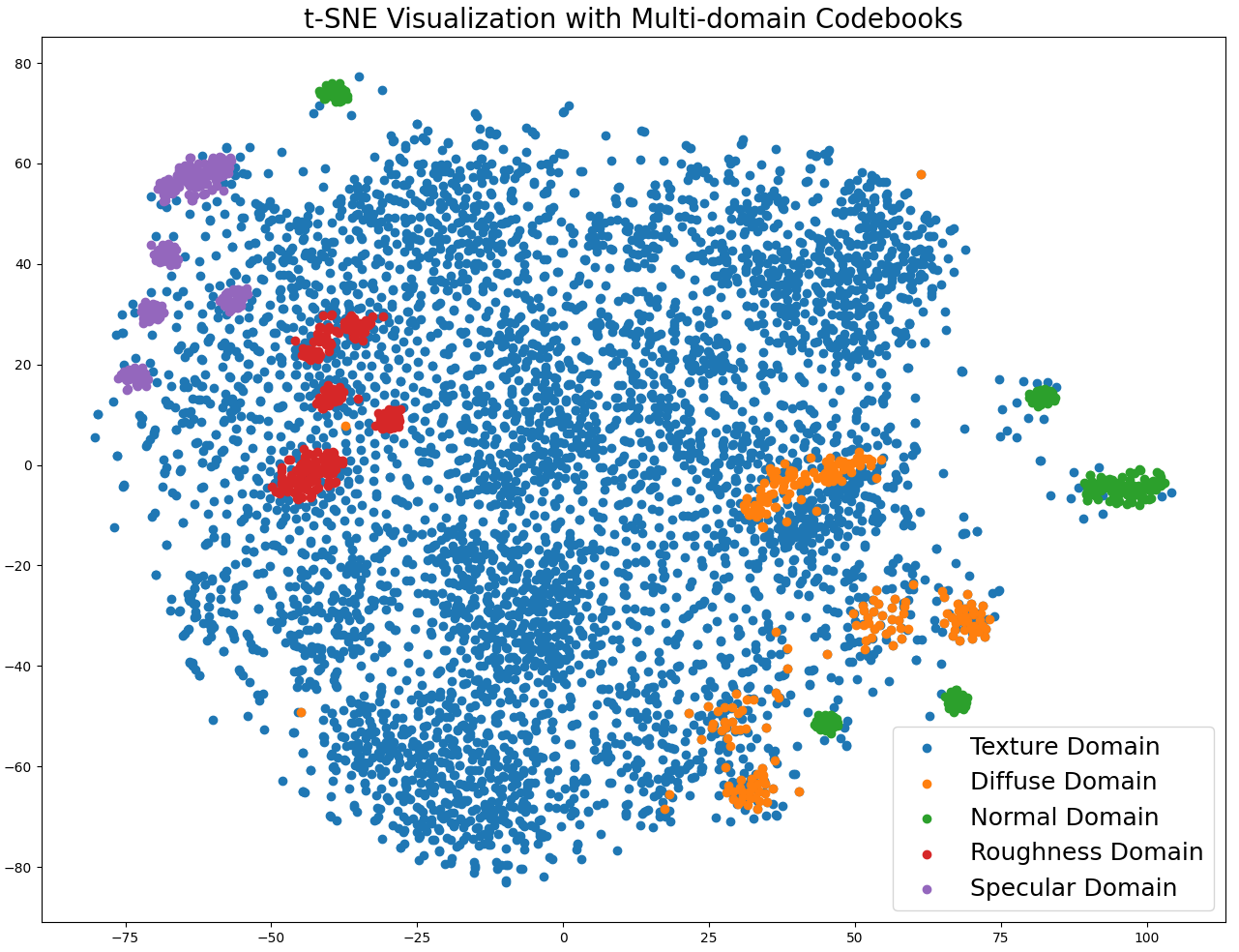}
    \caption{t-SNE distribution of latent feature $\hat{z}_q$ for reflectance data and RGB data.}
    \label{fig:supp-codebook}
\end{figure}

To further validate the distribution of different domain data, we extracted the discrete features $\hat{z}_q$ of the data and employed t-SNE to visualize them in 2D space. With this approach, we can visually assess the key structural features of different domain data. In this study, the t-SNE results~(shown in Fig.~\ref{fig:supp-codebook}) show that our domain-specific images are highly clustered internally and have good separability in the feature space.

\section{Details of Swapper}\label{swapper}
We conducted a detailed comparison of swapper module designs, as shown in Tab.~\ref{supptab:swapper}. 
We use a different face recognition model~\cite{wang2018cosface} as the ID vector extractor and Deep3D~\cite{Deng19} as a pose estimator.
Both ID-retrival and pose errors are evaluated by L2 distances. 
We observed that previous identity injection techniques (\eg, Faceshifter~\cite{li2020faceshifter} and SimSwap~\cite{chen2020simswap}) were unsuitable for our framework due to their low-quality results and convergence difficulties. This is because even minor perturbations could lead to significant changes in discrete codes, making it challenging to maintain facial structure integrity. It's important to note that identity injection should not involve all upsample layers, as this can lead the network to learn superfluous details at the expense of identity preservation.
In the training process of the swapper module, we use ProjectedGAN to maintain a good convergence speed which keeps consistent with other approaches. We observed that the pre-trained VQGAN~\cite{esser2021taming} discriminator would result in the swapper failing to converge, while the convergence speed would be too slow if trained from scratch.

\begin{figure*}[t]
    \centering
    \includegraphics[width=0.9\linewidth]{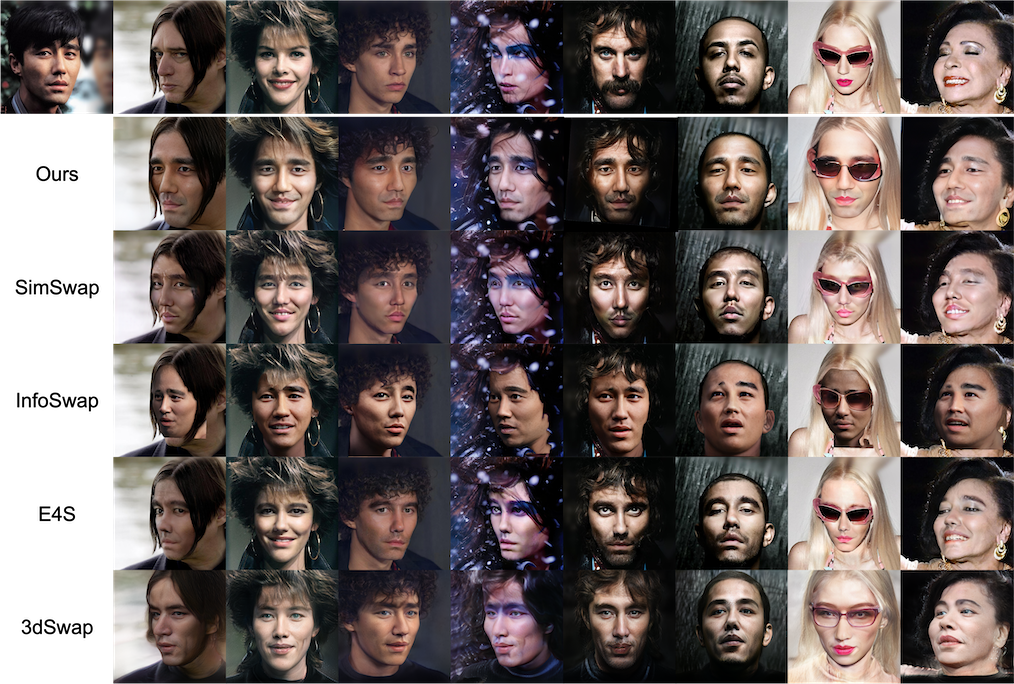}
    \caption{RGB-domain swapping comparison with other methods (\eg, SimSwap~\cite{chen2020simswap}, InfoSwap~\cite{gao2021inforswap}, E4S~\cite{liu2023fine}, and 3dSwap~\cite{li20233dfaceswapping}). Input source identity and target images are all from CelebAMaskHQ~\cite{CelebAMask-HQ} dataset.}
    \label{fig:supp-swapper-1}
\end{figure*}

\begin{figure*}[t]
    \centering
    \includegraphics[width=0.9\linewidth]{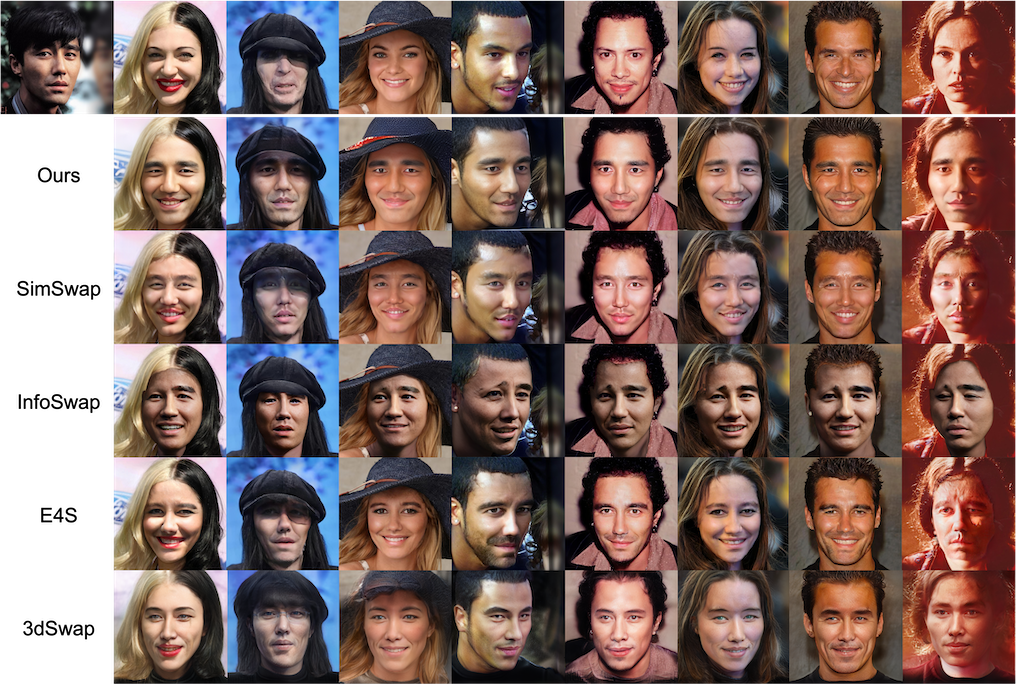}
    \caption{RGB-domain swapping comparison with other methods (\eg, SimSwap~\cite{chen2020simswap}, InfoSwap~\cite{gao2021inforswap}, E4S~\cite{liu2023fine}, and 3dSwap~\cite{li20233dfaceswapping}). Input source identity and target images are all from CelebAMaskHQ~\cite{CelebAMask-HQ} dataset.}
    \label{fig:supp-swapper-2}
\end{figure*}

\begin{figure*}[t]
    \centering
    \includegraphics[width=0.9\linewidth]{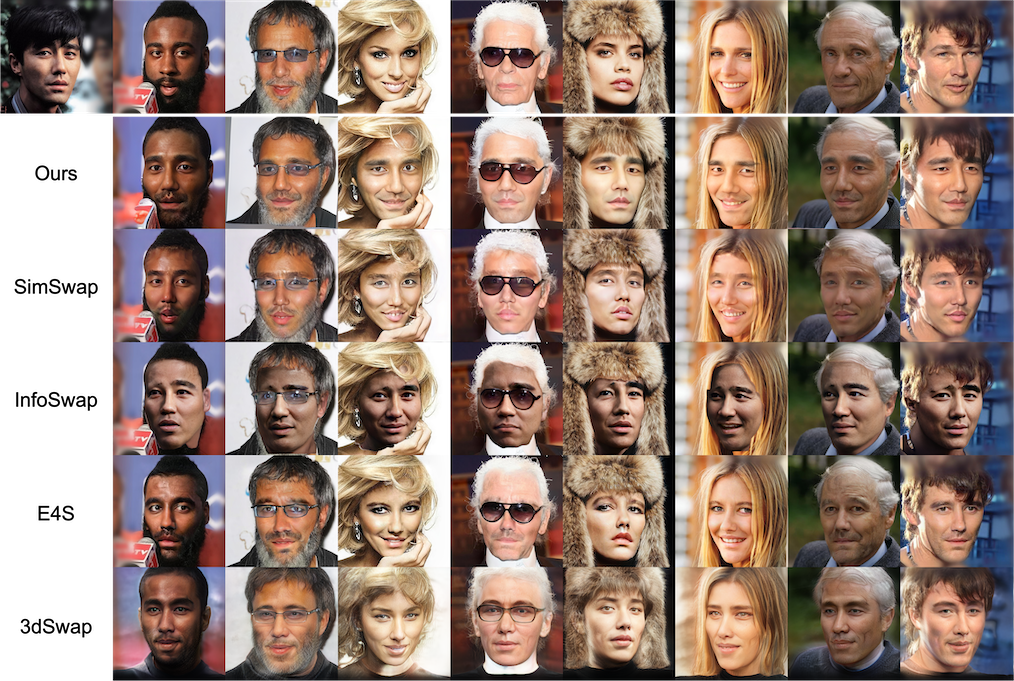}
    \caption{RGB-domain swapping comparison with other methods (\eg, SimSwap~\cite{chen2020simswap}, InfoSwap~\cite{gao2021inforswap}, E4S~\cite{liu2023fine}, and 3dSwap~\cite{li20233dfaceswapping}). Input source identity and target images are all from CelebAMaskHQ~\cite{CelebAMask-HQ} dataset.}
    \label{fig:supp-swapper-3}
\end{figure*}

In addition, we tested the results of face-swapping on RGB images, as shown in Fig.~\ref{fig:supp-swapper-1},~\ref{fig:supp-swapper-2} and~\ref{fig:supp-swapper-3}. It can be seen that our method can maintain good pose and identity in the RGB domain, and also achieves state-of-the-art swapping results in the RGB domain.
\begin{table}[t]
\centering
\resizebox{0.450\textwidth}{!}{
\begin{tabular}{@{}l|c|c@{}}
\toprule
\multicolumn{1}{c|}{Configuration} & ID-Retrival $\uparrow$  & pose $\downarrow$ ~ \\ \midrule
~F16                            & N/A & N/A ~\\
~F16 + F32                      & 0.894 & 0.0143 ~\\
~F16 + F32 + F64                & 0.941 & 0.0132 ~\\ 
~F16 + F32 + F64 + F128 + F256  & 0.933 & \textbf{0.0128} ~\\ \hline
~F16 + F32 + F64 + F128 (Ours)  & \textbf{0.965} & 0.0129 ~\\ \bottomrule
\end{tabular}}
\caption{Comparison of swapper module under different configurations. F16 means that the source identity is injected from an upsample layer with a feature size of $16\times16$.}
\label{supptab:swapper}
\end{table}

\section{UV Stitching}\label{stitching}
\begin{figure}[t]
    \centering
    \includegraphics[width=\linewidth]{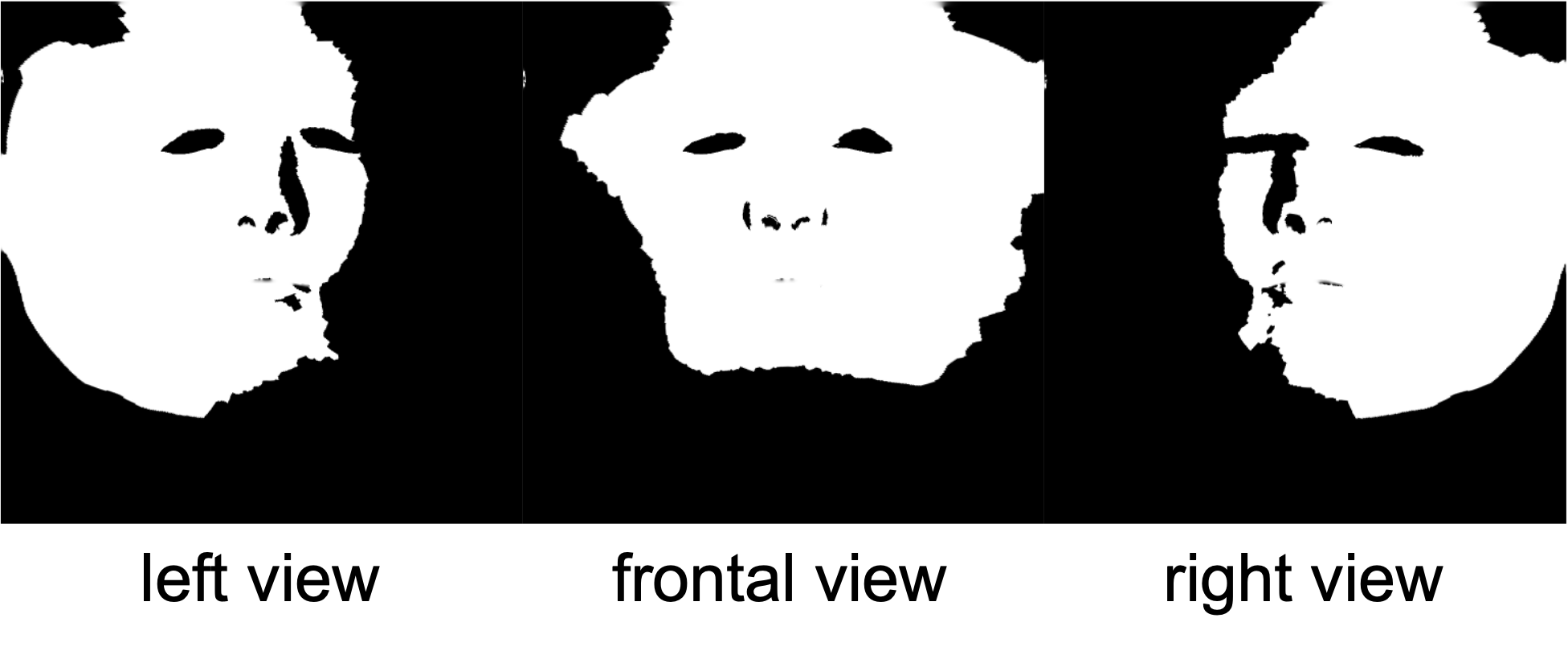}
    \caption{Example of three-view UV stitching masks.}
    \label{fig:supp-stitching}
\end{figure}

In our ID2Reflectance framework, we can pre-define three angles of view to ensure that the reflectance swapping is done consistently.
After tested on various templates, our swapping is robustly captured within an angle range of approximately -45 to +45 degrees. We further tested at ±30 degrees, finding a larger overlapping area in terms of effectiveness. Fig.~\ref{fig:supp-stitching} is a visualization of the mask used in our stitching process across different views.
The areas that are never covered by the mask will be filled by the template.
\begin{table}[t]
\centering
\resizebox{0.480\textwidth}{!}{
\begin{tabular}{@{}l|c|c|c|c@{}}
\toprule
\multicolumn{1}{c|}{Numbers} & Diffuse  & Specular  & Roughness & Normal ~ \\ \midrule
~30 subjects         & 25.09 & 24.12 & 24.78  & 23.55  ~\\
~60 subjects         & 28.63 & 27.58 & 28.22  & 27.05  ~\\ 
~90 subjects         & 30.54 & 29.84 & 30.41  & 29.22  ~\\ \midrule
~115 subjects & 31.62 & 30.96 & 31.59  & 30.32 ~\\ \bottomrule
\end{tabular}}
\caption{Comparison of ID2Reflectance framework under different training data. We calculate the average PSNR between the reconstructed and the ground truth reflectance maps on our test set.}
\label{supptab:ID2ReflectanceConfig}
\end{table}

\begin{table*}[t]
\centering
    \resizebox{0.9\linewidth}{!}{%
    \begin{tabular}{ l | c | c | c | c | cccccc }
    \toprule
    \multirow{2}{*}{Method} & \multirow{2}{*}{Avg. ITA $\downarrow$} & \multirow{2}{*}{Bias $\downarrow$} & \multirow{2}{*}{Score $\downarrow$} & \multirow{2}{*}{MAE $\downarrow$} & \multicolumn{6}{c}{ITA per skin type $\downarrow$} \\
    \cline{6-11}
    & & & & & I & II & III & IV & V & VI \\
    \midrule
    Deep3D~\cite{Deng19} & $22.57$ & $24.44$ & $47.02$   & $27.98$ & $\mathbf{8.92}$   & $\mathbf{9.08}$   & $8.15$    & $10.90$   & $28.48$   & $69.90$ \\
    GANFIT~\cite{Gecer19:ganfit}   & $62.29$ & $31.81$ & $94.11$   & $63.31$ & $94.80$   & $87.83$   & $76.25$   & $65.05$   & $38.24$   & $11.59$     \\
    MGCNet~\cite{Shang20}  & $21.41$ & $17.58$ & $38.99$   & $25.17$  & $19.98$   & $12.76$   & $8.53$   & $\mathbf{9.21}$   & $22.66$   & $55.34$    \\
    DECA~\cite{Feng2021} & $28.74$ & $29.24$ & $57.98$   & $38.17$  & $9.34$   & $11.66$   & $11.58$   & $16.69$   & $39.10$   & $84.06$   \\
    INORig~\cite{Bai21}  & $27.68$ & $28.18$ & $55.86$   & $33.20$   & $23.25$   & $11.88$   & $\mathbf{4.86}$    & $9.75$    & $35.78$   & $80.54$    \\
    CEST~\cite{Wen21} & $35.18$ & $12.14$ & $47.32$   & $29.92$  & $50.98$   & $38.77$   & $29.22$   & $23.62$   & $21.92$   & $46.57$     \\
    TRUST~\cite{feng2022towards} & $13.87$ & $\mathbf{2.79}$ & $\mathbf{16.67}$ & $\mathbf{18.41}$ & $11.90$   & $11.87$   & $11.20$   & $13.92$  & $\mathbf{16.15}$ & $18.21$ \\
    ID2Albedo~\cite{id2albedo} & $\mathbf{12.07}$ & $4.91$ & $16.98$ & $23.33$ & $18.30$   & $9.13$   & $5.83$   & $9.46$  & $19.09$ & $\textbf{10.59}$ \\
    \hline
    Ours & $14.21$ & $4.22$ & $18.43$ & $22.02$ & $12.91$   & $13.11$   & $9.68$   & $10.22$  & $17.72$ & $21.63$ \\
    \bottomrule
    \end{tabular}} %
    \caption{Comparison to state-of-the-arts on the FAIR benchmark~\cite{feng2022towards}. We utilize the FAIR official metrics, such as average ITA error, bias score (standard deviation), total score (avg. ITA+Bias), mean average error, and average ITA score per skin type in degrees (I: very light, VI: very dark).}
    \label{supptab:fair}
\end{table*}

\begin{figure*}[t]
    \centering
    \includegraphics[width=0.95\linewidth]{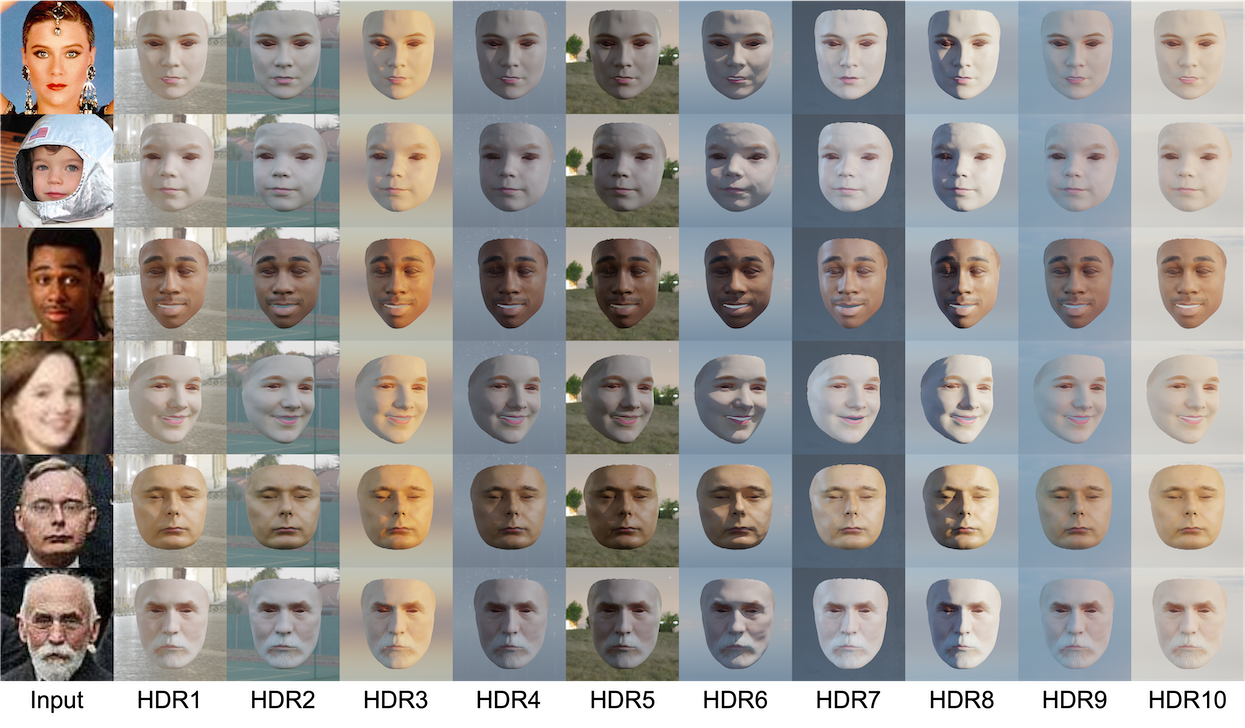}
    \caption{Relighting results for occluded and low-quality images. Each column of images maintains the same HDR illumination.}
    \label{fig:supp-vis-1}
\end{figure*}

\begin{figure*}[t]
    \centering
    \includegraphics[width=0.95\linewidth]{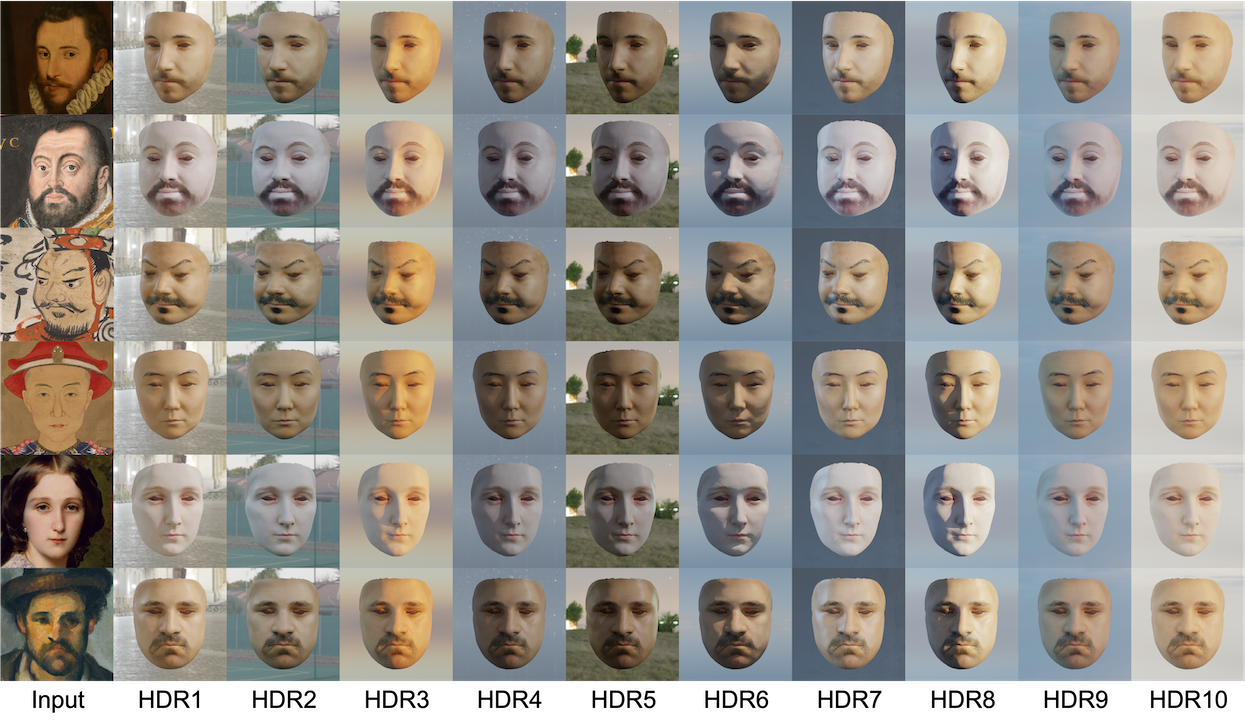}
    \caption{Relighting results for people in oil painting photographs. Each column of images maintains the same HDR illumination.}
    \label{fig:supp-vis-2}
\end{figure*}

\begin{figure*}[t]
    \centering
    \includegraphics[width=0.95\linewidth]{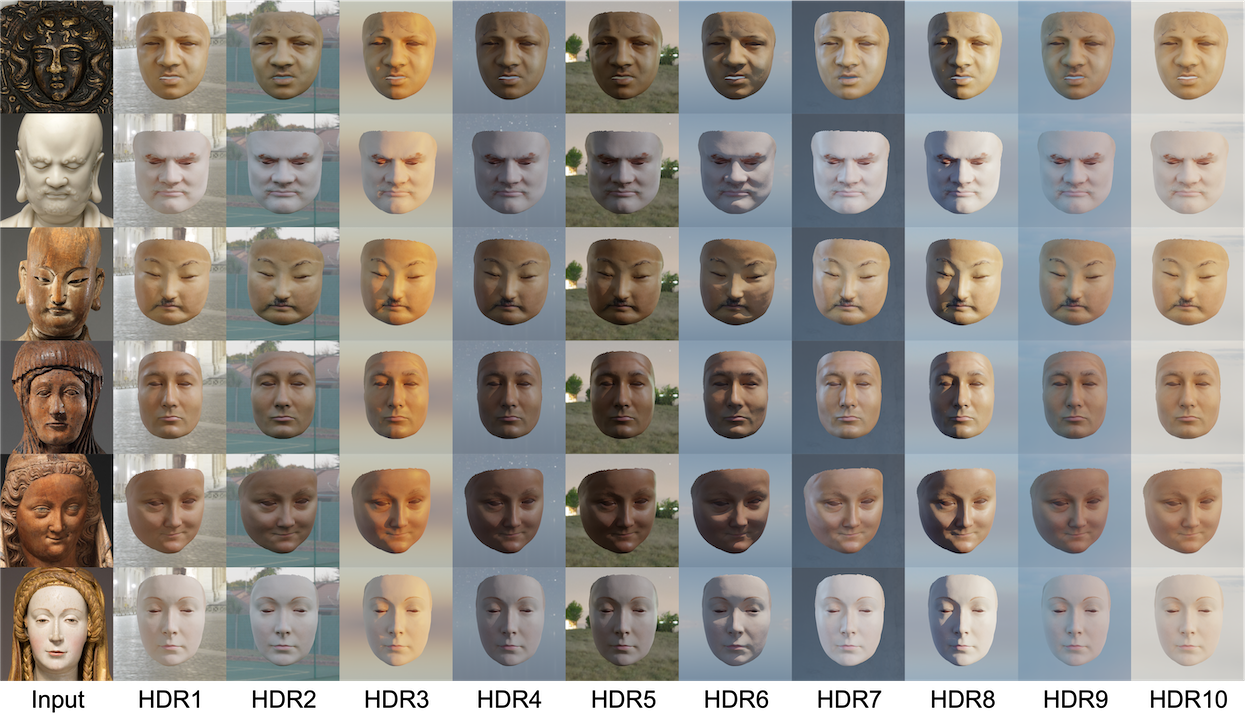}
    \caption{Relighting results for people from sculptures. Each column of images maintains the same HDR illumination.}
    \label{fig:supp-vis-3}
\end{figure*}
\section{More Qualitative Results}\label{vis}
We first verified the reconstruction ability of a multi-domain codebook under different numbers of training data. As shown in Tab.~\ref{supptab:ID2ReflectanceConfig}, the reconstruction quality varies with the amount of captured data. We observed that the more training data used, the better the final effect, but the improvement becomes slower.

Subsequently, we performed a comparison on the FAIR benchmark to verify the fairness of the method, as shown in Tab.~\ref{supptab:fair}. As can be seen, our method achieves comparable results in terms of ITA score, bias, and total results.

In addition, we show further results under challenging input conditions, as shown in Fig.~\ref{fig:supp-vis-1},~\ref{fig:supp-vis-2} and~\ref{fig:supp-vis-3}. Note that the distinguishing feature of our framework is the ability to predict fine reflectance for inputs such as low-quality, occlusion, or oil portraits. The reconstruction range can even be extended to any image from which facial features can be extracted, thus demonstrating excellent generalization. In contrast, traditional fitting and inpainting solutions that rely on the original image as the supervisory signal often have limited results with low-quality inputs.
\end{document}